\documentclass[10pt,twocolumn,letterpaper]{article}

\usepackage{cvpr}              

\usepackage{url}
\usepackage{amsfonts}       
\usepackage{nicefrac}       
\usepackage{microtype}      
\usepackage{xcolor}         
\usepackage{arydshln}
\usepackage[normalem]{ulem}
\usepackage{color}
\usepackage{graphicx}
\usepackage{amsmath,bm}
\usepackage{amssymb}
\usepackage{array}
\usepackage{comment}
\usepackage{booktabs}
\usepackage{caption}
\usepackage{colortbl} 
\usepackage{multirow}
\usepackage{makecell}
\usepackage{wrapfig}
\usepackage{pifont}
\usepackage{tabularray}
\usepackage{paralist}
\usepackage{adjustbox}
\usepackage{rotating}
\usepackage{enumerate}
\usepackage{CJKutf8}
\usepackage{algorithm}
\usepackage{algpseudocode}

\usepackage{tcolorbox}
\usepackage{adjustbox}
\usepackage{pgfplots}
\pgfplotsset{compat=1.17}
\tcbuselibrary{breakable}
\newcommand{\customfont}{\linespread{0.7}\selectfont}

\definecolor{myblue}{RGB}{52,218,247}
\definecolor{myred}{RGB}{255,90,90}
\definecolor{mypink}{RGB}{239,43,159}
\definecolor{myupdate}{RGB}{254,243,222}
\definecolor{myfrozen}{RGB}{237,255,255}
\definecolor{ired}{RGB}{229,72,72}
\definecolor{igreen}{RGB}{80,219,144}
\definecolor{nmblue}{RGB}{216,234,247}
\definecolor{lightgreen}{RGB}{196,250,222}
\definecolor{lightblue}{RGB}{211,227,253}
\definecolor{lightgrey}{RGB}{201,203,209}
\definecolor{linkred}{RGB}{255,0,49}
\definecolor{textred}{RGB}{255,242,234}
\definecolor{imageblue}{RGB}{224,250,255}
\definecolor{videogreen}{RGB}{214,250,232}
\definecolor{vgreen}{RGB}{39,190,110}
\definecolor{linkcolor}{RGB}{255,0,0}
\definecolor{urlcolor}{RGB}{255,105,180}
\definecolor{citepcolor}{RGB}{66,168,235}

\usepackage{tikz}

\newcommand{\circlenum}[1]{%
    \resizebox{!}{0.8em}{%
        \tikz[baseline=(char.base)]{
            \node[shape=circle, fill=black, inner sep=0.8pt, text=white] (char) {#1};
        }%
    }%
}

\usepackage[misc,geometry]{ifsym} 
\definecolor{cvprblue}{rgb}{0.21,0.49,0.74}
\usepackage[pagebackref,breaklinks,colorlinks,allcolors=cvprblue]{hyperref}

\makeatletter
\DeclareRobustCommand\onedot{\futurelet\@let@token\@onedot}
\def\@onedot{\ifx\@let@token.\else.\null\fi\xspace}
\def\@fnsymbol#1{\ensuremath{\ifcase#1\or *\or
   \mathsection\or \mathparagraph\or \|\or **\or \dagger\dagger
   \or \ddagger\ddagger \else\@ctrerr\fi}}
\makeatother

\begin{document}

\def\paperID{3506} 
\def\confName{CVPR}
\def\confYear{2025}

\title{Universal Scene Graph Generation}

\author{
Shengqiong Wu, \quad Hao Fei\footnote{The corresponding author.}, \quad Tat-seng Chua \\
National University of Singapore \\
{\tt\small swu@u.nus.edu, haofei37@nus.edu.sg, dcscts@nus.edu.sg}
}


\twocolumn[{%
\renewcommand\twocolumn[1][]{#1}%
\maketitle
\vspace{-30pt}
\begin{center}
\centering
\includegraphics[width=0.99\linewidth]{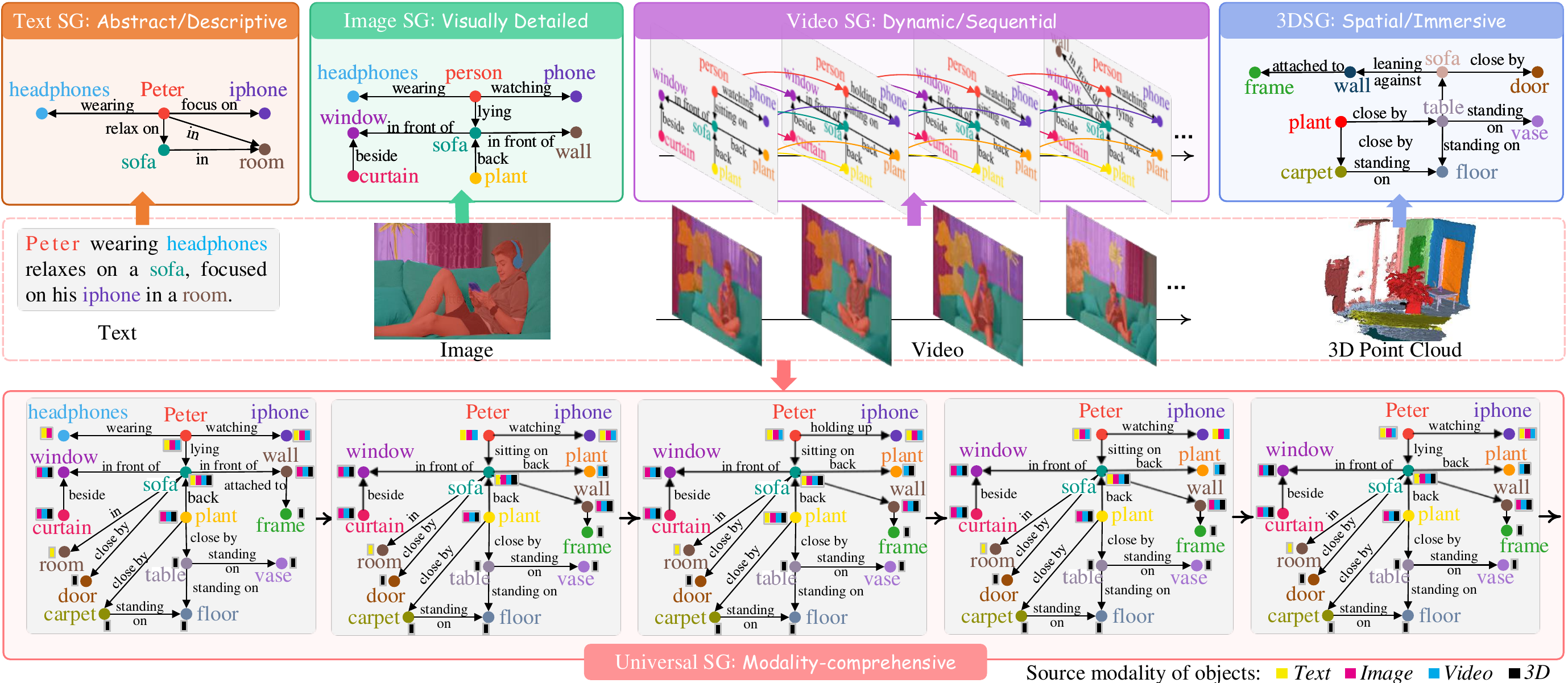}
\vspace{-5pt}
\captionof{figure}{
Illustrations of SGs (top) of single modalities in text, image, video, and 3D, and our proposed Universal SG (bottom).
Note that the USG instance shown here is under the combination of four complete modalities, while practically any modality can be absent freely.
Also, the temporal coreference edges are omitted for visual clarity (a full version is given in the Appendix).
}
\label{fig:example-intro}
\end{center}
}]
\def\thefootnote{*}\footnotetext{Corresponding author.}
\def\thefootnote{\arabic{footnote}}

\begin{abstract}
Scene graph (SG) representations can neatly and efficiently describe scene semantics, which has driven sustained intensive research in SG generation.
In the real world, multiple modalities often coexist, with different types, such as images, text, video, and 3D data, expressing distinct characteristics.
Unfortunately, current SG research is largely confined to single-modality scene modeling, preventing the full utilization of the complementary strengths of different modality SG representations in depicting holistic scene semantics.
To this end, we introduce Universal SG (USG), a novel representation capable of fully characterizing comprehensive semantic scenes from any given combination of modality inputs, encompassing modality-invariant and modality-specific scenes.
Further, we tailor a niche-targeting USG parser, USG-Par, which effectively addresses two key bottlenecks of cross-modal object alignment and out-of-domain challenges.
We design the USG-Par with modular architecture for end-to-end USG generation, in which we devise an object associator to relieve the modality gap for cross-modal object alignment.
Further, we propose a text-centric scene contrasting learning mechanism to mitigate domain imbalances by aligning multimodal objects and relations with textual SGs. 
Through extensive experiments, we demonstrate that USG offers a stronger capability for expressing scene semantics than standalone SGs, and also that our USG-Par achieves higher efficacy and performance.
The project page is \url{https://sqwu.top/USG/}.
\end{abstract}

\vspace{-5mm}
\section{Introduction}
\label{sec:intro}

\vspace{-2mm}
Scene understanding is a fundamental topic in computer vision and artificial intelligence, aiming to comprehend and interpret scenes in a manner akin to human perception.
Within scene understanding, SG generation \cite{VG-KrishnaZGJHKCKL17,Motifs-ZellersYTC18,Pair-Net-10634834} stands as a pivotal task, seeking to identify and classify all constituent objects in a given scene, along with their attributes and interrelationships.
This process constructs a semantic graph representation that facilitates a comprehensive understanding of the specific scene. 
SGs are widely applied in various real-world applications \cite{abs-2304-03696,PhamHLS24,abs-2405-15321,abs-2408-06926,fei2024dysen,0001W0ZZLH24}, such as autonomous driving \cite{abs-2403-19098,GreveBVBV24,WangMZYF24}, robot navigation \cite{SeymourMCS022,abs-2403-17846}, augmented reality \cite{TaharaSNI20,KalkofenMS09}, etc.
Consequently, SG generation has garnered significant research attention in past decades \cite{BCTR-abs-2407-18715,VCTree-TangZWLL19,SGTR-LiZ022,VRD-LuKBL16,RelTR-CongYR23,STTran-CongLARY21}.

Humans perceive the world through a multitude of sensory modalities, acquiring information via different channels to form a complete perception of their environment. 
Thus, beyond images, scenes can be represented through various other modalities as well, including text, video, and 3D formats. 
Correspondingly, current research has focused on constructing scene graphs across these different modalities, such as generations of Image SG (ISG) \cite{VG-KrishnaZGJHKCKL17,Motifs-ZellersYTC18,VCTree-TangZWLL19}, Textual SG (TSG) \cite{AMR-SG-T5-choi2022scene,FACTUAL-LiCZQHLJT23}, Video SG (VSG) \cite{VidVRD-shang2017video, VidOR-shang2019annotating,STTran-CongLARY21}, and 3D SG (3DSG) \cite{3DDSG-WaldDNT20,VL-SAT-WangCZ0TS23,SGFN-WuWTNT21}. 
Moreover, due to the inherent nature of each modality, SGs in each modality possess distinct capabilities, as illustrated in Fig. \ref{fig:example-intro}. 
Specifically:

\begin{compactitem}
    \item \textbf{ISG} – Images provide concrete visual details, enabling SGs to precisely describe object locations, sizes, and visual attributes, but typically cannot directly convey temporal sequences or dynamic changes.

    \item \textbf{TSG} – Compared to the concreteness of vision, text can flexibly describe abstract entities, actions, events, and abstract relationships between objects and entities that are not visually explicit, but usually lack specific visual and spatial details.

    \item \textbf{VSG} – Building upon static images, videos are more adept at expressing dynamic events, actions, and temporal changes.

    \item \textbf{3DSG} On top of 2D vision, 3D scenes can further model objects and their spatial relationships, sizes, orientations, and other 3-dimensional spatial attributes.

\end{compactitem}

\noindent It is evident that SG representations across different modalities provide complementary insights for describing a whole scene semantics. 
Conversely, this implies that relying solely on one single modality's SG cannot offer a comprehensive scene representation. 
In practical applications, an ideal process involves users providing inputs in any single modality or even multiple combinations and the system simultaneously extracting modality-specific and modality-shared scene information to derive a unified SG representation for a comprehensive scene understanding. 
Unfortunately, current research \cite{RelTR-CongYR23,OED-WangLCL24,CCL-3DDSG-ChenWLLWH24,SGTR-LiZ022} communities study SGs for different modalities separately, and a universal representation encompassing all modalities does not yet exist. 
To bridge this gap, this paper proposes a Universal SG (USG) representation capable of characterizing any combination of modalities. 
As shown in Fig. \ref{fig:example-intro}, USG can incorporate all the characteristics of individual SGs, capturing a comprehensive semantic scene from any given combination of input modalities.

Yet, realizing such a universal SG generation presents two non-trivial challenges from the methodological perspective. 
First, regarding \textbf{model architecture}, the current community has most largely explored singleton-modality SG generation methods \cite{Motifs-ZellersYTC18,PSG-YangAGZZ022,EGTR-ImNPLP24,TPT-ZhangPYHMC24,3DDSG-WaldDNT20}. 
To achieve USG parsing, a direct approach would be to use a pipeline paradigm, i.e., first parsing each modality's SG separately, and then merging the individual SGs into one USG \cite{PhamHLS24,abs-2406-19255}. 
The most challenging issue here is merging identical objects across multiple modalities while retaining modality-specific scene information. 
However, because the independent SG parsers for certain modalities operate in isolation, it leads to critical problems: 
i), complementary semantic information across modalities may be overlooked; 
ii), due to differences in feature spaces across modalities, it becomes difficult to precisely align identical objects across modalities, resulting in a final USG that is neither concise nor effective.
Second, regarding \textbf{data}, the lack of annotated USG data is a significant hurdle. 
Since manual annotation is labor-intensive, a feasible solution is to leverage the existing abundance of single-modality SG annotation datasets to learn USG, i.e., through joint training on various singleton SG datasets. 
Unfortunately, significant domain divergence exists among different modality data. 
For example, 3DSG data might focus solely on static indoor scenes, VSGs are mostly biased towards action-rich scenes, while only TSGs hold for general domains. 
Consequently, the resulting USG parser inevitably suffers from scene biases, thereby limiting its effectiveness.

To address these challenges, we present a USG Parser (termed \textbf{USG-Par}), capable of end-to-end scene parsing from any modality inputs, outputting a USG representation. 
Technically, USG-Par works sequentially through 5 main modules:
Step \circlenum{1}, modality-specific encoders encode inputs from different modalities; 
Step \circlenum{2}, we employ Mask2Former \cite{mask2former-ChengMSKG22} as a shared mask decoder to generate representations for scene objects; 
Step \circlenum{3}, an object associator is devised to determine whether objects from different modalities are identical. Specifically, to eliminate the modality gap, objects are transformed into their respective modality-specific feature spaces before object association and alignment; 
Step \circlenum{4}, a relation proposal constructor generates the most feasible relation pairs by modeling object-level interactions; 
Step \circlenum{5}, a relation decoder finally predicts the final relations among different objects based on the selected pairs from the previous step.
During model training, to combat the issue of data domain imbalance, we propose a \emph{text-centric scene contrasting learning} mechanism, where considering that the text modality is scene-unbiased in the general domain, we align objects and relations from various modalities to the objects and relations in the text space of TSG.

\begin{figure*}[!t]
    \centering
    \includegraphics[width=0.99\textwidth]{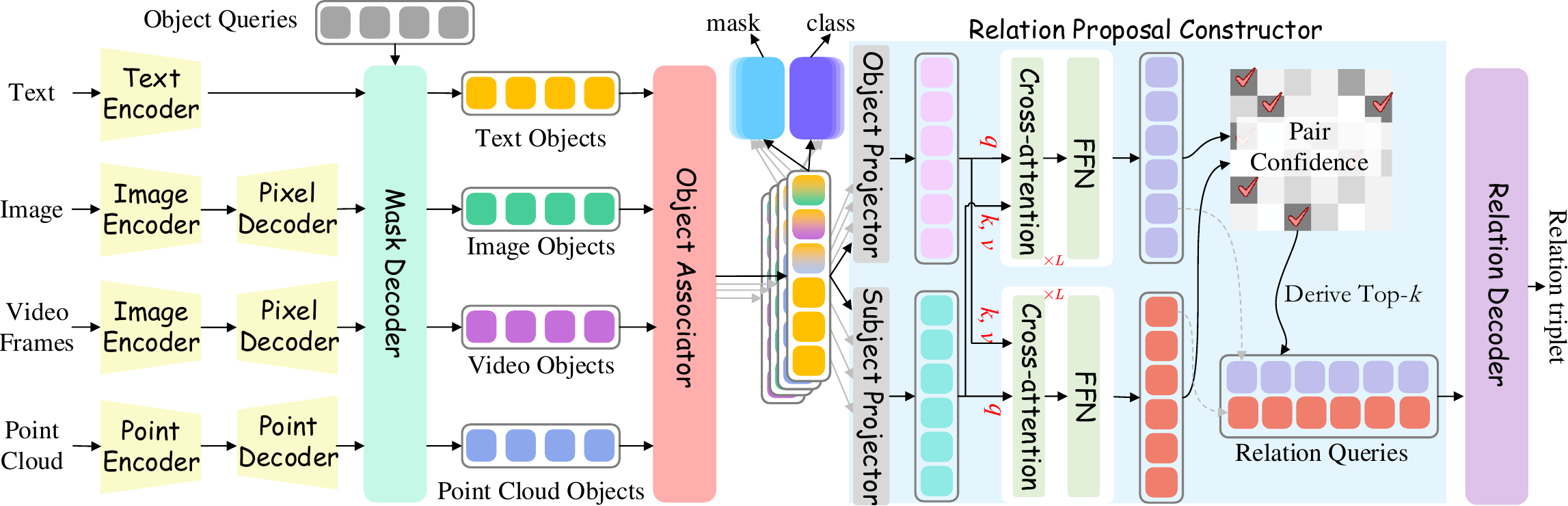}
    \vspace{-2mm}
    \caption{Overview of USG-Par architecture.
    It mainly consists of five modules, including modality-specific encoders, shared mask decoder, object associator, relation proposal constructor, and relation decoder.
    }
    \label{fig:frame}
    \vspace{-5mm}
\end{figure*}

Experimental results demonstrate that USG provides a more powerful and comprehensive scene representation compared to standalone SGs of individual modalities. 
Furthermore, extensive experimental results on various SG benchmark datasets indicate that, 
i), the proposed USG-Par achieves significant performance improvements in single-modality scene graph parsing; 
and ii), in multi-modality SG parsing, USG-Par accurately constructs associations between objects from different modalities, achieving better performance than pipeline approaches.
Further, we show that USG-Par effectively handles scene parsing for unseen scene domains and unseen modality combinations, thanks to the text-centric scene contrasting learning for weak supervision.

In summary, this work makes two primary contributions to the community.
\textbf{First}, we are the first to propose a Universal SG representation for holistic semantic scene understanding.
\textbf{Second}, we introduce a novel USG generator, USG-Par, which effectively addresses cross-modal object alignment and out-of-domain challenges simultaneously.

\vspace{-2mm}
\section{Universal Scene Graph}
\label{sec:formatting}
\vspace{-1mm}

\subsection{Preliminary}
\vspace{-2mm}

The concept of the SG is initially introduced in \cite{JohnsonKSLSBL15} as a visually grounded graph structure representing object instances within an image, where edges depict pairwise relationships between these objects. 
Formally, given an image $\mathcal{I} \in \mathbb{R}^{H \times W}$, ISG generation task is defined as:
\setlength\abovedisplayskip{3pt}
\setlength\belowdisplayskip{3pt}
\begin{equation}
    F(\mathcal{G}^\mathcal{I}|\mathcal{I}) = F(\{\mathcal{O}, \mathcal{R}\}|\mathcal{I}),
\end{equation}
where $\mathcal{O}$ is the set of objects, with each object node $o_i$ characterized by a bounding box or a mask segmentation $m_i$ and an associated category label $c_i^{o} \in \mathbb{C}^{\mathcal{O}}$, \ie $o_i = (c_i^{o}, m_i)$. 
$\mathcal{R}$ is the set of relations, with each directed edge $r_{i,j}$ between two objects (subject and object) described by a predicate $c_{i,j}^{r}  \in \mathbb{C}^{\mathcal{R}}$, i.e., $r_{i,j} = (o_i, c_{i,j}^{r}, o_j )$.
$\mathbb{C}^{\mathcal{O}}$ and $\mathbb{C}^{\mathcal{R}}$ means the object and predicate classes.

Recently, by incorporating temporal dimensions, VSG \cite{VG-KrishnaZGJHKCKL17} generation has been introduced:
\setlength\abovedisplayskip{4pt}
\setlength\belowdisplayskip{4pt}
\begin{equation}
    F(\mathcal{G}^\mathcal{V}|\mathcal{V}) = F(\{\mathcal{G}_t^{\mathcal{V}}\}_{t=1}^{T}|\mathcal{V}) = F(\{\mathcal{O}_t, \mathcal{R}_t\}_{t=1}^T|\mathcal{V}),
\end{equation}
where $ \mathcal{V} \in \mathbb{R}^{T \times H \times W}$ is the input video,  and $\mathcal{G}_t^{\mathcal{V}} = \{\mathcal{O}_t, \mathcal{R}_t\}$ denotes a ISG at $t$-th frame.

To enable spatially immersive understanding, methods for generating 3DSG \cite{3DDSG-WaldDNT20} are then developed:
\setlength\abovedisplayskip{3pt}
\setlength\belowdisplayskip{3pt}
\begin{equation}
    F(\mathcal{G}^\mathcal{D}|\mathcal{D}) = F(\{\mathcal{O}, \mathcal{R}\}|\mathcal{D}),
\end{equation}
where $\mathcal{D} \in \mathbb{R}^{P \times 6}$ is the input 3D point clouds, with $P$ standing for the number of point clouds of interest and 6 representing xyz coordinates plus RGB values.
Similar to ISG, each 3D object $o_i$ in the object set $\mathcal{O}$ is identified by an instance segmentation $m_i \in \{0,1\}^{P}$ and a category label $c_i^{o}$

Lastly, text, as a highly abstract and flexible modality for scene description, has motivated research on generating SGs from textual inputs \cite{SchusterKCFM15}, formulated as:
\begin{equation}
    F(\mathcal{G}^\mathcal{S}|\mathcal{S}) = F(\{\mathcal{O}, \mathcal{R}\}|\mathcal{S}),
\end{equation}
where $\mathcal{S} \in \mathbb{R}^{L}$ is a given text.
In TSGs, each object node is defined solely by its category label.

\vspace{-1mm}
\subsection{USG Definition}
\vspace{-1mm}

In contrast to existing methods that focus solely on single-modality SG generation, we define our USG generation task as being able to handle any single-modality SG generation as well as any combination of modalities.
Given a set of input data in various modalities (e.g., image $\mathcal{I}$, video $\mathcal{V}$, 3D point cloud $\mathcal{D}$ and text $\mathcal{S}$), USG generation is formulated as:
\begin{equation}
\begin{aligned}
    & F(\mathcal{G}^{\mathcal{U}}|\{\mathcal{I}, \mathcal{V}, \mathcal{D}, \mathcal{S}\}) = F(\{\mathcal{O}, \mathcal{R}\}|\{\mathcal{I}, \mathcal{V}, \mathcal{D}, \mathcal{S}\}), \\
\end{aligned}
\end{equation}
where $\mathcal{O} = \{\mathcal{O}^{*}\}$, $* \in \{\mathcal{I}, \mathcal{V}, \mathcal{D}, \mathcal{S} \}$ represents the set of objects across all modalities. 
Each node involves a category label $c_i^o \in \mathbb{C}^{\mathcal{O}}$ and a segmentation mask $m_i$. 
Additionally, for objects extracted from textual descriptions, we construct a positional binary mask to indicate their locations.
$\mathcal{R} = \{\mathcal{R}^{*}, \mathcal{R}^{* \times \diamond}\}$, $*, \diamond \in \{\mathcal{I}, \mathcal{V}, \mathcal{D}, \mathcal{S} \} $ \text{and} $ * \ne \diamond $. $\mathcal{R}^{*}$ includes both intra-modality relationships and inter-modality associations $\mathcal{R}^{* \times \diamond}$. 
For example, in Fig. \ref{fig:example-intro}, the text and image describe the same scene; thus, the textual object ``\textit{Peter}'' in the TSG should correspond to the visual object ``\textit{person}'' in the ISG. 
Similarly, the ``\textit{sofa}'' in the 3DSG corresponds to the object ``\textit{sofa}'' in the ISG.

\vspace{-1mm}
\section{Methodology}
\vspace{-1mm}
Our model consists of five main modules, as shown in Fig. \ref{fig:frame}. 
\textbf{First}, we first extract the modality-specific features with a modality-specific backbone. 
\textbf{Second}, we employ a shared mask decoder to extract object queries for various modalities. 
These object queries are then fed into the modality-specific object detection head to obtain the category label and tracked positions of the corresponding objects. 
\textbf{Third}, the object queries are input into the object associator, which determines the association relationships between objects across modalities. 
\textbf{Fourth}, a relation proposal constructor is utilized to retrieve the most confidential subject-object pairs. 
\textbf{Finally}, a relation decoder is employed to decode the final predicate prediction between the subjects and objects.

\vspace{-1mm}
\subsection{Modality-specific Encoder}
\vspace{-1mm}

To encode each modality, we propose using specialized encoders for each one:
\textbf{1) Text Encoder.} We employ OpenCLIP \cite{Radford2021LearningTV} to encode the input text $\mathcal{S}$ to obtain the text contextualized features, $\bm{H}^{\mathcal{S}}$.
\textbf{2) Image Encoder and Pixel Decoder.} We adopt the frozen CLIP-ConvNeXt \cite{liu2022convnet} as the backbone image encoder to model the given image/video inputs, yielding frozen feature $\bar{\bm{H}}^{\mathcal{I}/\mathcal{V}}$.  
The pixel decoder, adapted from Mask2Former, consists of multi-stage deformable attention layers that transform the frozen features $\bar{\bm{H}}^{\mathcal{I}/\mathcal{V}}$, into the fused multi-scale feature $\{\bm{H}^{\mathcal{I}/\mathcal{V}}_i\}^3_{i=1}$, with the same channel dimension, where $i$ is the layer index, and $i = 3$ corresponds to the highest-resolution feature.
\textbf{3) Point Encoder and Point Decoder.} We employ Point-BERT \cite{Point-BERT-YuTR00L22} as the point encoder to encode the input point cloud $\mathcal{D}$, generating the super-point features, $\bar{\bm{H}}^{\mathcal{D}}$. 
The point decoder is designed to propagate the super-point features to each point hierarchically, producing multi-scale point features $\{\bm{H}^{\mathcal{D}}\}_{i=1}^3$, where $i=3$ denotes point clouds features with the original number points.
All features are projected into a common $d$-dimensional space using a linear layer.

\begin{figure}[!t]
  \centering
   \includegraphics[width=0.98\linewidth]{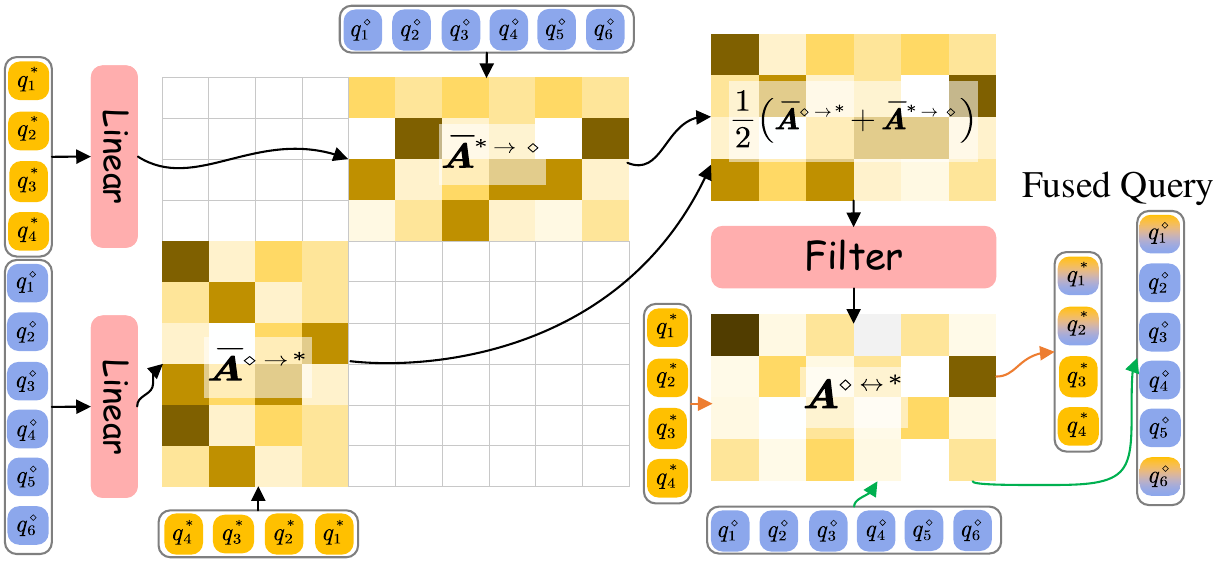}
   \vspace{-2mm}
   \caption{Illustration of the object associator for establishing associations between different modalities.}
   \label{fig:obj_ass}
   \vspace{-6mm}
\end{figure}

\vspace{-1mm}
\subsection{Shared Mask Decoder}
\vspace{-1mm}

Objects across different modalities can provide complementary information, facilitating cross-modal learning. 
Therefore, we employ a shared mask decoder framework to implicitly integrate these cross-modal complementary features.
Following \cite{mask2former-ChengMSKG22}, we utilize multi-scale features and a cascaded decoder to perform masked cross-attention between modality-specific features $\bm{H}^{*}$ and the corresponding object query features $\bm{X}^{*}_l \in \mathbb{R}^{N_q^{*} \times d}, * \in \{\mathcal{I},\mathcal{V},\mathcal{D},\mathcal{S}\}$ as follows:
\setlength\abovedisplayskip{3pt}
\setlength\belowdisplayskip{3pt}
\begin{equation}
    \bm{X}^{*}_l = \text{softmax}(\bm{M}^{*}_{l-1} + \bm{Q}_{l-1}^{*}{\bm{K}_{l-1}^{*\top}}) \bm{V}_{l-1}^{*} + \bm{X}_{l-1}^{*},
\end{equation}
where $N_q^{*}$ is the number of queries and $l$ is the layer index.
$\bm{M}^{*}_{l-1}$ is the binarized output of the resized mask prediction from the previous stage.
$\bm{X}_0^{*}$ denotes input object query features to the mask decoder.
$\bm{Q}_{l-1}^{*} = F_q(\bm{X}_{l-1}^{*})$, while $\bm{K}_{l-1}^{*} = F_k(\bm{H}^{*}) $ and $\bm{V}_{l-1}^{*} = F_v(\bm{H}^{*})$.
Here, $F_q(\cdot)$, $F_k(\cdot)$ and
$F_v(\cdot)$ are linear transformations as typically applied in attention mechanisms.
In practice, for image, video, and 3D data, $\bm{H}^{*}$ is sampled from the multi-scale feature output $\{\bm{H}^{\mathcal{I}/\mathcal{V}/\mathcal{D}}\}_{i=1}^{3}$, while for text, we employ $\bm{H}^{\mathcal{S}}$ across different scales.
In addition, for video data, to effectively capture the temporal information across frames, we incorporate a transformer-based temporal encoder $F_{temp}$ to model the temporal relationships between objects.
After $L^{mask}$ layers, we obtain the refined object queries $\bm{Q}^{*} = \{\bm{q}_i^{*}\}_{i=1}^{N_q^*}$.

\vspace{-1mm}
\subsection{Object Associator}
\vspace{-2mm}

The biggest challenge in USG generation would be accurately merging identical objects across multiple modalities.
To establish robust associations and bridge the modality gap, we propose projecting objects into each other's feature spaces using a transformation layer before determining association relationships:
\setlength\abovedisplayskip{4pt}
\setlength\belowdisplayskip{4pt}
\begin{equation}
   \begin{aligned}
   \bar{\bm{A}}^{* \rightarrow \diamond} &= \text{cos}(F_{* \rightarrow \diamond}(\bm{Q}^{*}), \bm{Q}^{\diamond}), \\
   \bar{\bm{A}}^{\diamond \rightarrow *} &= \text{cos}(F_{\diamond \rightarrow * }(\bm{Q}^{\diamond}), \bm{Q}^{*}),\\
   \bar{\bm{A}}^{* \leftrightarrow \diamond} &= (\bar{\bm{A}}^{* \rightarrow \diamond} + \bar{\bm{A}}^{ \diamond \rightarrow *}) / 2 \; , \\
   \end{aligned}
\end{equation}
where $F_{* \rightarrow \diamond}(\cdot)$ is a linear transformation of $\bm{Q}^{*}$, similar to $F_{\diamond \rightarrow *}(\cdot)$.
We then design a filtering module to further refine and learn feasible sparse association pairs. In practice, we employ a CNN-based architecture, which leverages local details while efficiently filtering out redundant noise.
The output is a refined association matrix, denoted as $\bm{A}^{* \leftrightarrow \diamond}$.
In Fig. \ref{fig:obj_ass} we illustrate the above process in detail.

\vspace{-2mm}
\subsection{Modality-specific Object Detection Head}
\vspace{-1mm}

Detecting objects involves predicting the segmentation mask and category label (containing a ``no object'' label)  from each object query.
To achieve simultaneous object detection across different modalities while retaining modality-specific scene information, we employ modality-specific heads. 
The design is consistent across modalities; here, we illustrate the process in the image modality as an example.
Upon establishing association relationships between objects across modalities, we fuse the object query via $\bm{q}^{\mathcal{I}}_i= \bm{q}^{\mathcal{I}}_i + \sum_j \bm{A}_{i,j}^{\mathcal{I} \leftrightarrow *}(\bm{q}^{*}_j), * \in \{\mathcal{V}, \mathcal{D}, \mathcal{S}\}$, allowing the incorporation of complementary information from other modalities to enrich the object embeddings of the current modality.
Then, a category classifier is applied on the fused query features $\bm{q}_{i}^{\mathcal{I}}$ to yield category label probability predictions $\bar{c}_{i}^{o, \mathcal{I}}$ for each segment.
For mask prediction, each binary mask prediction $\bar{m}_i^{\mathcal{I}} \in [0,1]^{H \times W}$ is obtained by computing the dot product between the fused query features and per-pixel embeddings $\bm{H}_3^{\mathcal{I}}$, i.e., $\bar{m}_i^{\mathcal{I}} = \text{sigmoid}(\text{MLP}(\bm{Q}^{\mathcal{I}}) \cdot \bm{H}_3^{\mathcal{I}\top})$.

\vspace{-2mm}
\subsection{Relation Proposal Constructor}
\vspace{-1mm}

We employ a subject and object projector, implemented as an MLP, to generate subject and object embeddings $\bm{E}^{obj}, \bm{E}^{sub}$, respectively.
we omit the modality superscript for simplicity.
A straightforward approach would involve calculating relationship embeddings by combining embeddings for all possible subject-object pairs and subsequently classifying the relationship predicates. 
However, such an exhaustive pairwise computation is computationally infeasible. 
Moreover, intuitively, improving recall of relevant pairs correlates with enhanced relationship recall, suggesting that focusing on the most promising object pairs could increase computational efficiency and overall performance.
To this end, we introduce a Relation Proposal Constructor (RPC) to selectively identify promising object pairs, as shown in Fig. \ref{fig:frame}. 
Specifically, we design a two-way relation-aware cross-attention mechanism $F_{\text{CA}}(q, k, v)$, to iteratively refine subject and object features as follows:
\begin{equation}
\begin{aligned}
   \bm{X}^{sub}_{l} &= F_{\text{CA}}^{obj \rightarrow sub}(\bm{X}^{sub}_{l-1},\bm{X}^{ obj}_{l-1}, \bm{X}^{obj}_{l-1}), \\
   \bm{X}^{obj}_{l} &= F_{\text{CA}}^{sub \rightarrow obj}(\bm{X}^{ obj}_{l-1}, \bm{X}^{sub}_{l-1}, \bm{X}^{sub}_{l-1)}, \\
\end{aligned}
\end{equation}
where $l$ denotes the layer index, and $\bm{X}^{sub}_{0} = \bm{E}^{sub}, \bm{X}^{obj}_{0} = \bm{E}^{obj} $. 
Following $L^{RPC}$ layers of interaction, we compute the cosine similarity between the refined subject and object embeddings, resulting in a Pair Confidence Matrix $\bm{C} = \text{cos}( \bm{X}^{sub}_{L}, \bm{X}^{obj}_{L})$.
We then perform a top-$k$ selection on the $\bm{C}$, with the top-$k$ indices used to retrieve the corresponding subject and object queries from $\bm{X}^{sub}_{L}$,$\bm{X}^{obj}_{L}$, which are denoted as $\bm{Q}^{sub}$ and $\bm{Q}^{obj}$ respectively.

\vspace{-2mm}
\subsection{Relation Detector}
\vspace{-1mm}
After retrieving the potential subject and object queries from RPC, they are concatenated together along the length dimension to construct relationship queries:
\begin{equation}
    \bm{Q}^{rel} = [\bm{Q}^{sub}+\bm{E}^{sub};\bm{Q}^{obj} +\bm{E}^{obj}],
\end{equation}
where $[;]$ means the concatenation. 
Then, we employ a transformer-based relation decoder applying cross-attention with keys and values from contextualized input features and self-attention on queries to predict the final relation:
\begin{equation}
   \bm{X}^{rel}_{l} = \text{F}_{CA}^{rel}(\bm{X}^{rel}_{l-1}, \bm{H}, \bm{H}),
\end{equation}
where $\bm{X}^{rel}_{0} = \bm{Q}^{rel}$ is the relationship query features into the relation decoder.
Since relationship analysis primarily focuses on examining semantic information, we leverage the contextualized representations generated by each modality-specific encoder as $\bm{H}$.
After applying $L^{rel}$ transformer layer, we apply a relationship classifier on the refined relationship queries $\bm{X}^{rel}_{L}$ to predict the final relationship probabilities.

\begin{figure}[t]
  \centering
   \includegraphics[width=0.98\linewidth]{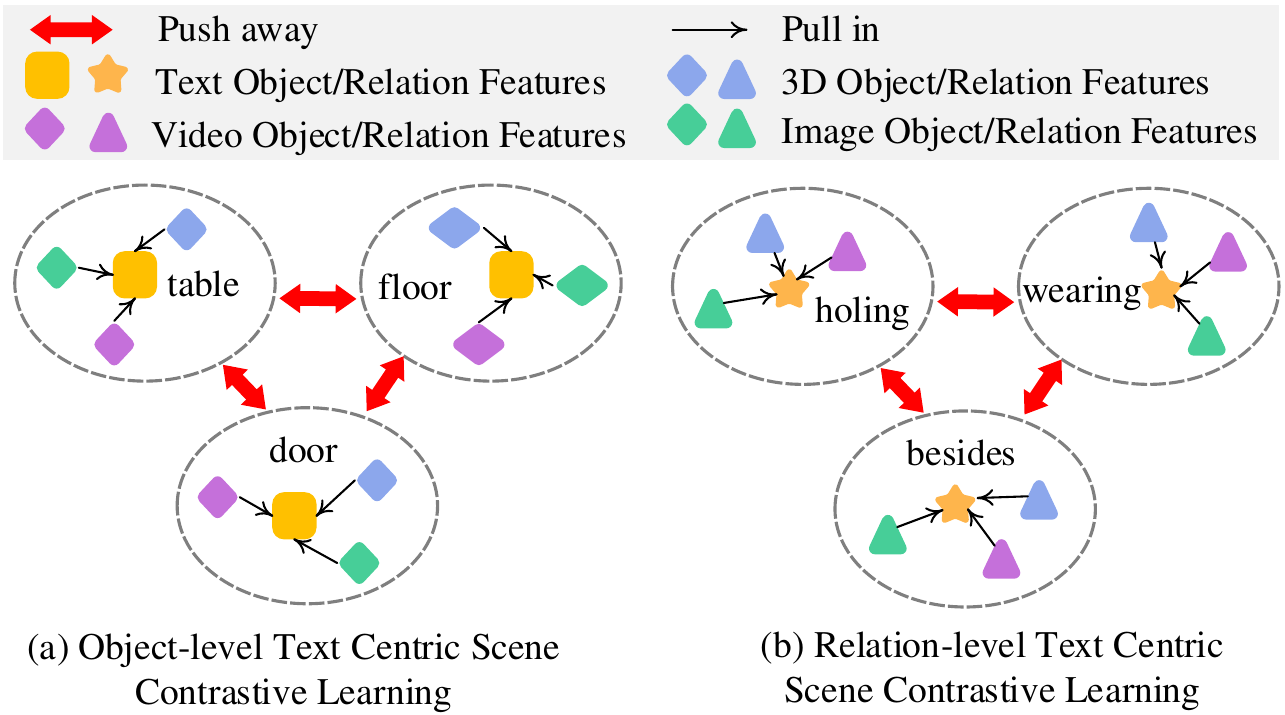}
   \vspace{-2mm}
   \caption{
   Illustration of the object-level and relation-level text-centric scene contrasting learning mechanism.
   }
   \label{fig:text_cons}
   \vspace{-5mm}
\end{figure}

\vspace{-1mm}
\subsection{Training with Domain-balancing Strategy}
\vspace{-1mm}

This section elaborates on the training objectives and strategies to optimize our system.

\vspace{-4mm}
\paragraph{Object Detection Loss.}
During training, we first apply Hungarian matching between the predicted and ground-truth entity masks to assign object queries to entities in text, video, image, and 3D modalities.
This assignment is then used to supervise the mask predictions and category label classifications.
We employ a sigmoid Cross-Entropy (CE) loss $L_{cls}^{o}$ as in \cite{MindererGSNWDMA22}, computed between the ground truth object classes and logits obtained by computing the inner product of the object queries with the text embeddings of the category names.
Moreover, following \cite{Pair-Net-10634834}, a binary CE loss $L_{ce}$ and Dice loss $L_{dice}$ are leveraged for segmentation.
\begin{equation}
    \mathcal{L}_{obj} =  \lambda_{cls} L_{cls}^{o} + \lambda_{ce}L_{ce} + \lambda_{dice}L_{dice},
\end{equation}
where $\lambda_{cls}, \lambda_{ce}, \lambda_{dice}$ are the parameter weights.

\vspace{-4mm}
\paragraph{Object Association Loss.}
To optimize the object associator, we take the ground-truth association matrix, which is a binary matrix, as the supervised signal.
Due to the sparsity of the association matrix, we utilize a weighted binary CE loss, $\mathcal{L}_{ass}$, to ensure stable training by significantly increasing the weight of the positive entries in the matrix.

\vspace{-4mm}
\paragraph{Relation Classification Loss.}
For relation predicate classification, we employ a sigmoid CE loss $\mathcal{L}_{cls}^{r}$ for the predicate classification, similar to object category classification. 
In addition, to supervise the relation proposal constructor in selecting the most confidential object pairs for further relation classification, we utilize a weighted binary CE loss, $\mathcal{L}_{pair}$, on the pair confidence matrix $\bm{C}$:
\begin{equation}
\begin{aligned}
    \mathcal{L}_{rel} &= \mathcal{L}_{cls}^{r} + \mathcal{L}_{pair}. \\
\end{aligned}
\end{equation}

\vspace{-4mm}
\paragraph{Text-centric Scene Contrastive Learning.}

A significant challenge when leveraging single-modal SG data for USG-Par learning is the domain imbalance across modalities, compounded by the lack of USG data for various modality combinations, which can result in suboptimal performance.
To address these, we propose a text-centric scene contrastive learning approach that aligns other modalities with text data, attributed with two key advantages: \textbf{1)} TSG data encompass the most diverse and general domain, and \textbf{2)} binding information from other modalities into text effectively addresses the scarcity of USG data for certain modality combinations \cite{imagebind-GirdharELSAJM23,languagebind-ZhuLNYCWPJZLZ0024}.
Considering the unique characteristics of SG, we design both object-level and relation-level text-centric contrastive learning, as illustrated in Fig. \ref{fig:text_cons}.
Given text-* pairs as inputs, where * represents image, video, or 3D modalities, we extract textual queries $\bm{Q}^{\mathcal{T}}$ and other modal queries $\bm{Q}^{*}, * \in \{\mathcal{I}, \mathcal{V}, \mathcal{D}\}$. 
Positive targets are constructed when corresponding textual objects are present in other modalities; otherwise, they serve as negative targets.
Thus, the object-level text-centric contrastive loss is formulated as:
\begin{equation}
    \mathcal{L}_{cons}^{o} = - \sum_{y^+} \log \frac{\exp(\mathbf{x} \cdot \mathbf{y}^+)}{\exp(\mathbf{x} \cdot \mathbf{y}^+) + \sum_{y^-} \exp(\mathbf{x} \cdot \mathbf{y}^-)},
\end{equation}
where $ \mathbf{x} $, $ \mathbf{y}^+ $, and $ \mathbf{y}^- $ are query embeddings of text-* pairs, their positive targets, and negative targets, sampled from object queries $\bm{Q}^{\mathcal{T}}$ and $\bm{Q}^{*}$, respectively,
Similarly, we compute the relation-level text-centric contrastive loss $\mathcal{L}_{cons}^{r}$.
The total contrastive loss is given by $\mathcal{L}_{cons} = \mathcal{L}_{cons}^{o} + \mathcal{L}_{cons}^{r}$.

\vspace{-4mm}
\paragraph{Training Target in Total.}
We combine all four loss terms in a linear manner as our final loss function:
\begin{equation}
    \mathcal{L} = \alpha \mathcal{L}_{obj} + \beta \mathcal{L}_{ass} + \gamma \mathcal{L}_{rel} + \eta \mathcal{L}_{cons},
\end{equation}
where $\alpha, \beta$, $\gamma$ and $\eta$ are the weights for the loss terms.

\begin{table}[!t]
  \centering
\fontsize{8}{9}\selectfont
\setlength{\tabcolsep}{1.2mm}
\begin{tabular}{llcccc}
\hline
 \bf Modality & \bf Dataset & \bf \#Obj. & \bf \#Rela.  & \bf \#Tri. & \bf \#Ins.\\
\hline
Text & FACTUAL~\cite{FACTUAL-LiCZQHLJT23} & 4,042 & 1,607 & 40,149 & 40,369 \\
\cdashline{1-6}
\multirow{2}{*}{Image} & VG~\cite{VG-KrishnaZGJHKCKL17} & 5,996 &  1,024 & 1,683,231 & 108,077 \\
& PSG~\cite{PSG-YangAGZZ022} & 133 & 56 & 275,371 & 48,749 \\
\cdashline{1-6}
\multirow{2}{*}{Video} & AG~\cite{AG-JiK0N20} & 36 &	25 &	772,013 &	288,782 \\
& PVSG~\cite{PVSG-YangPLGCL0ZZLL23} & 126 &	65 &	4,587 &	400 \\
\cdashline{1-6}
3D & 3DDSG~\cite{3DDSG-WaldDNT20} & 528 &	39 &	543,956 &	1,335 \\
\hline
\multirow{5}{*}{{Multimodal}} & $\mathcal{S}-\mathcal{I}$ & 6,089 & 1,235 & 1,791,309 & 124,357  \\
& $\mathcal{S}-\mathcal{V}$ & 150 & 132 & 6,751 & 400  \\
& $\mathcal{S}-\mathcal{D}$ & 724 & 257 & 230,865 & 46,173 \\
& $\mathcal{I}-\mathcal{V}$ & 126 & 65 & 4,587 & 400 \\
& $\mathcal{I}-\mathcal{D}$ & 345 & 75 & 7,689 & 4,492 \\
\hline
\end{tabular}%
\vspace{-2mm}
\caption{
Statistics of SG datasets. `\#Obj.' and `\#Rela.' denote the number of the object and relation categories, respectively. `\#Tri.' is relation triplets count, and `\#Ins.' is instance count. 
}
\vspace{-5mm}
\label{tab:data}
\end{table}

\vspace{-2mm}
\section{Experimental Settings}
\vspace{-1mm}
\paragraph{Datasets and Resources.}

We conduct experiments on two distinct groups of datasets to comprehensively evaluate our method's SG generation capability in single and multiple modalities.
\textbf{1) Single modality}, we employ a range of well-established datasets: VG~\cite{VG-KrishnaZGJHKCKL17} and PSG~\cite{PSG-YangAGZZ022} for images, AG~\cite{AG-JiK0N20} and PVSG\cite{PVSG-YangPLGCL0ZZLL23} for videos, 3DDSG~\cite{3DDSG-WaldDNT20} for 3D scenes, and FACTUAL~\cite{FACTUAL-LiCZQHLJT23} for text-based scenes.
Notably, some of these datasets provide only bounding box annotations; thus, we use SAM-2~\cite{abs-2408-00714} to generate pseudo-segmentation masks, using the bounding boxes as prompts.
\textbf{2) Multiple modalities}, on the one hand, we construct multimodal SGs by leveraging paired text-image/video/3D data. 
We utilize GPT-4o \cite{openai2023gpt4} to parse initial TSGs from textual captions, and then link textual and visual objects through label matching.
To increase the diversity and richness of textual descriptions in these multimodal pairs, we rephrase and enrich captions, allowing for flexible and partially non-literal associations with visual content.
Additionally, for image-video cross-modal SGs, we pair randomly selected frames with temporally non-adjacent video segments. Similarly, we pair 2D image views with corresponding 3D scenes for image-3D cross-modal SGs. 
Tab. \ref{tab:data} shows the statistics of the datasets, and we provide further details on data construction in the Appendix.

\begin{table}[!t]
  \centering
\fontsize{8}{9}\selectfont
\setlength{\tabcolsep}{1.4mm}
\begin{tabular}{lcccc}
\hline
\bf Method & \bf Backbone  & \bf R/mR@20 & \bf R/mR@50  & \bf R/mR@100 \\
\hline
IMP~\cite{IMP-XuZCF17} & R50 & 16.5 / 6.5  & 18.2 / 7.1&  18.6 / 7.2   \\
Motifs~\cite{Motifs-ZellersYTC18} & R50   & 20.0 / 9.1 & 21.5 / 9.6  & 22.0 / 9.7  \\
VCTree~\cite{VCTree-TangZWLL19} & R50 & 20.6 / 9.7   & 22.1 / 10.2 & 22.5 / 10.2  \\
GPS-Net~\cite{GPS-Net-LinDZT20} & R50 & 17.8 / 7.0  & 19.6 / 7.5  & 20.1 / 7.7  \\
PSGTR~\cite{PSG-YangAGZZ022} & R50 &   28.4 / 16.6   & 34.4 / 20.8   & 36.3 / 22.1  \\
PSFormer~\cite{PSG-YangAGZZ022} & R50 &  18.1 / 14.8   & 19.6 / 20.1  & 17.4 / 18.7 \\
HiLo~\cite{HiLo-0002SC23} & R50 & \underline{34.1} / 23.7    & 40.7 / 30.3 & 43.0 / 33.1  \\
DSGG~\cite{DSGG-Hayder024} & R50 &  32.7 / \underline{30.8}   & \underline{42.8} / \underline{38.8} & \underline{50.0} / \underline{43.4}  \\
Pair-Net~\cite{Pair-Net-10634834}  & R50 & 29.6 / 24.7  & 35.6 / 28.5   & 39.6 / 30.6   \\
Pair-Net~\cite{Pair-Net-10634834}  & Swin-B &  33.3 / 25.4  &  39.3 / 28.2 &  42.4 / 29.7  \\
\cdashline{1-5}
\rowcolor{lightgreen} USG-Par$^\natural$ (Ours) & OpenCLIP  &  35.7 /   29.9  &	  44.6 /   40.9  &   51.3 /   42.7  \\
\rowcolor{lightgreen} USG-Par (Ours) & OpenCLIP  & \bf 36.9 /  \bf 32.1  &	 \bf 46.4 /  \bf 41.7  &  \bf 52.4 /  \bf 44.6  \\
\hline
\end{tabular}
\vspace{-2mm}
\caption{
Evaluation on the PSG~\cite{PSG-YangAGZZ022} under the SGDet task.
$^\natural$ means the model is trained solely on the corresponding single-modality dataset, here, PSG.
The top baseline results are underlined, and the best overall performance is highlighted in bold.
The tables below follow the same format.
}
\vspace{-2mm}
\label{tab:psg}
\end{table}

\begin{table}[!t]
  \centering
\fontsize{8}{9}\selectfont
\setlength{\tabcolsep}{2.2mm}
\begin{tabular}{lccc}
\hline
{\bf Method} &  \bf R/mR@20 & \bf R/mR@50 & \bf R/mR@100 \\
\hline
IPS+T+1D Conv.~\cite{PVSG-YangPLGCL0ZZLL23}  & 2.79 / 1.24 & 2.80 / 1.47 & 3.10 / 1.59 \\
IPS+T+Trans.~\cite{PVSG-YangPLGCL0ZZLL23}  & \underline{4.02} / \underline{1.75} & \underline{4.41} / \underline{1.86} & \underline{4.88} / \underline{2.03} \\
VPS+1D Conv.~\cite{PVSG-YangPLGCL0ZZLL23}  & 0.60 / 0.27 & 0.73 / 0.28 & 0.76 / 0.29 \\
VPS+Trans.~\cite{PVSG-YangPLGCL0ZZLL23}  & 0.75 / 0.36 & 0.91 / 0.39 & 0.94 / 0.40 \\
\cdashline{1-4}
\rowcolor{lightgreen} USG-Par$^\natural$ (Ours) &  4.68 /  2.01 &	 5.37 /  2.02 &  6.15 / 3.03  \\
\rowcolor{lightgreen} USG-Par (Ours) & \bf 5.08 / \bf 2.23 &	\bf 6.64 / \bf 2.36 & \bf 7.45 / \bf 3.76  \\
\hline
\end{tabular}
\vspace{-2mm}
\caption{
Evaluation on the PVSG~\cite{PVSG-YangPLGCL0ZZLL23}.
}
\vspace{-6mm}
\label{tab:pvsg}
\end{table}

\vspace{-5mm}
\paragraph{Evaluation Metrics.}

We evaluate our methods following three standard evaluation tasks: 
1) predicate classification (\textbf{PreCls}); 2) scene graph classification (\textbf{SGCls}); 3) scene graph detection (\textbf{SGDet}).
Following previous work \cite{Pair-Net-10634834}, we adopt Recall@K (\textbf{R@$K$}) and mean Recall@K (\textbf{mR@$K$}) as evaluation metrics to measure the fraction of ground truth hit in the top $K$ predictions, where $K \in \{10, 20, 50, 100\}$.
For the TSG, we compute the \texttt{Set Match} and \texttt{SPICE} as in \cite{FACTUAL-LiCZQHLJT23} for evaluation.

\vspace{-6mm}
\paragraph{Implementation.}

We initialize the text and image encoders using OpenCLIP~\cite{Radford2021LearningTV}. 
Following the approach in \cite{mask2former-ChengMSKG22,LiY0DWZLCL24}, we design the pixel decoder. 
For the point encoder, we adopt Point-BERT~\cite{Point-BERT-YuTR00L22} as the initialization, and for the point decoder, inspired by \cite{Point-BERT-YuTR00L22}, we implement a hierarchical propagation strategy with distance-based interpolation. 
The mask decoder follows the design in ~\cite{mask2former-ChengMSKG22}.
We set the number of predefined learnable queries to 100.
The object associator is implemented as a 3-layer CNN with a kernel size of $3 \times 3$. 
The relation decoder comprises a 6-layer transformer with an embedding dimension of 256.
During training, we used the AdamW optimizer with an initial learning rate of $10e-4$.
More implementation details can refer to the Appendix.

\vspace{-2mm}
\section{Results and Analyses}
\vspace{-1mm}

\subsection{Main Observations}
\vspace{-1mm}
We compare USG-Par with the existing methods on single and multiple modalities data.

\begin{table}[!t]
  \centering
\fontsize{8}{9}\selectfont
\setlength{\tabcolsep}{4mm}
\begin{tabular}{lcc}
\hline
\multirow{2}{*}{\bf Method} & {\bf SGCls} & {\bf PreCls} \\
\cmidrule(r){2-2} \cmidrule(r){3-3}
& \bf R@20/50/100 & \bf R@20/50/100 \\
\hline
SGPN~\cite{SGPN-WaldDNT20}  & 27.0 / 28.8 / 29.0 & 51.9 / 58.0 / 58.5 \\
SGFN~\cite{SGFN-WuWTNT21}  & 27.5 / 29.2 / 29.2 & 52.6 / 58.9 / 59.4 \\
EdgeGCN~\cite{EdgeGCN-ZhangY0021}  & 28.0 / 29.8 / 29.8 & 54.7 / 60.9 / 61.5 \\
KISGP~\cite{KISG-ZhangLHQ21}  & 28.5 / 30.0 / 30.1 & 59.3 / 65.0 / 65.3 \\
\citeauthor{MKA-SG-FengH0WGM23}~\cite{MKA-SG-FengH0WGM23}  & - / 31.5 / 31.6 & - / 31.5 / 31.6 \\
VL-SAT~\cite{VL-SAT-WangCZ0TS23}  & 32.0 / 33.5 / 33.7 & 67.8 / 79.9 / 80.8 \\
CCL-3DDSG~\cite{CCL-3DDSG-ChenWLLWH24}  & \underline{37.6} / \underline{40.3} / \underline{45.7}  & \underline{73.6} / \underline{80.5} / \underline{82.9} \\
\cdashline{1-3}
\rowcolor{lightgreen} USG-Par$^\natural$ (Ours) &  36.6 /  41.4 /  46.2 & 71.9 /  81.0  /  83.4 \\
\rowcolor{lightgreen} USG-Par (Ours) &  \bf 37.9 / \bf 43.1 / \bf 46.9 & \bf 73.5 / \bf 81.7  / \bf 84.1 \\
\hline
\end{tabular}
\vspace{-2mm}
\caption{
Evaluation results on the 3DDSG~\cite{3DDSG-WaldDNT20} dataset. 
}
\vspace{-3mm}
\label{tab:3ddsg}
\end{table}

\begin{table}[!t]
  \centering
\fontsize{8}{9}\selectfont
\setlength{\tabcolsep}{1.5mm}
\begin{tabular}{lcccc}
\hline
\multirow{2}{*}{\bf Method} & \multicolumn{2}{c}{\bf Random} & \multicolumn{2}{c}{\bf Length} \\
\cmidrule(r){2-3} \cmidrule(r){4-5}
& \bf Set Match & \bf SPICE & \bf Set Match & \bf SPICE \\
\hline
SPICE-Parser~\cite{SPICE-AndersonFJG16}    & 13.00 & 56.15 & 0.94 & 38.04 \\
AMR-SG-T5~\cite{AMR-SG-T5-choi2022scene}        & 28.45 & 64.82 & 12.16 & 51.71 \\
CDP-T5~\cite{AMR-SG-T5-choi2022scene}         & 46.15 & 73.56 & 26.50 & 61.21 \\
VG-T5~\cite{VG-T5-SharifzadehBSST22}            & 11.54 & 47.46 & 2.94 & 42.98 \\
FACTUAL-T5 (pre)~\cite{FACTUAL-LiCZQHLJT23} & \underline{79.77} & \textbf{\underline{92.91}} & \underline{42.35} & \underline{82.43} \\
FACTUAL-T5~\cite{FACTUAL-LiCZQHLJT23}   & 79.44 & 92.23 & 38.65 & 80.76 \\
\cdashline{1-3}
\rowcolor{lightgreen} USG-Par$^\natural$ (Ours) &  80.40 & 87.53 & 39.75 & 83.69 \\
\rowcolor{lightgreen} USG-Par (Ours) & \bf 82.40 & 88.12 & \textbf{43.83} & \bf 84.38 \\
\hline
\end{tabular}
\vspace{-2mm}
\caption{
Performance on the FACTUAL~\cite{FACTUAL-LiCZQHLJT23} dataset.
}
\vspace{-2mm}
\label{tab:factual}
\end{table}

\begin{table}[!t]
  \centering
\fontsize{8}{9}\selectfont
\setlength{\tabcolsep}{1.0mm}
\begin{tabular}{lccccc}
\hline
\bf Method & $\mathcal{S}-\mathcal{I}$  & $\mathcal{S}-\mathcal{V}$ & $\mathcal{S}-\mathcal{D}$  & $\mathcal{I}-\mathcal{D}$ & $\mathcal{I}-\mathcal{V}$ \\
\hline
 & 75.4 / 25.4 & 73.3 / 1.9   & 71.1 / 13.3 &  39.1 / 12.6   & 35.4 / 4.2  \\
 & \cite{FACTUAL-LiCZQHLJT23} + \cite{Pair-Net-10634834} & \cite{FACTUAL-LiCZQHLJT23} + \cite{PVSG-YangPLGCL0ZZLL23} & \cite{FACTUAL-LiCZQHLJT23} + \cite{CCL-3DDSG-ChenWLLWH24} & \cite{Pair-Net-10634834} + \cite{PVSG-YangPLGCL0ZZLL23} & \cite{Pair-Net-10634834} + \cite{CCL-3DDSG-ChenWLLWH24} \\
\cdashline{1-6}
USG-Par$^\flat$ 
& 78.6 / 26.2 & 76.4 / 2.0  & 74.9 / 15.4 &  40.4 / 13.7   & 37.6 / 4.2 \\
USG-Par  
& \multirow{2}{*}{79.6 / 29.2} & \multirow{2}{*}{79.4 / 2.2}  & \multirow{2}{*}{77.9 / 17.4} &  \multirow{2}{*}{43.3 / 16.7}   & \multirow{2}{*}{41.6 / 7.2} \\
\quad - $\mathcal{L}_{cons}$ &  &  &  &  & \\
\rowcolor{lightgreen} USG-Par & 80.3 / 32.4  &  80.6 / 2.4   &	  79.2 / 20.2  &   47.3 / 18.6   &  42.7 / 9.5 \\
\hline
\end{tabular}
\vspace{-2mm}
\caption{
For SGDet task evaluation on multimodal inputs, we apply separate SG parsers per modality as referenced.
$^\flat$ means raining on corresponding multimodal data only, and 
$- \mathcal{L}_{cons}$ denotes training without text-centric scene contrastive loss.
We separately report the \texttt{Set Match} and \texttt{mR@50} scores for text and other modalities.
}
\vspace{-5mm}
\label{tab:multimodal}
\end{table}

\vspace{-5mm}
\paragraph{1) USG Generation in Single Modality.}
We present the experimental results for both single-dataset training and joint training across multiple datasets, as shown in Tab. \ref{tab:psg} \ref{tab:pvsg} \ref{tab:3ddsg} \ref{tab:factual}.
Additional results on other datasets are provided in Appendix.
In the single-dataset training setting, our model achieves comparable performance to the best-performing baselines and even slightly surpasses them on certain datasets, such as PVSG.
This demonstrates the effectiveness of our model design across different modalities, highlighting its capacity to enhance performance on individual datasets.
When comparing our joint-training results with baselines, our model consistently outperforms across all datasets. 
For example, on PSG, our method shows an average R@K score improvement of 3.2, indicating that joint learning effectively leverages single-modal SG data to boost overall performance.

\begin{figure}[!t]
    \centering
    \includegraphics[width=0.85\linewidth]{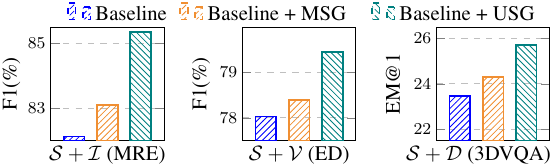}
    \vspace{-3mm}
    \caption{Comparison of multimodal tasks with and without USG integration. Baselines: multimodal relation extraction (MRE) \cite{Wu00BC23}, emotion detection (ED) \cite{10.1145/3689646}, and 3D visual QA (3DVQA) \cite{AzumaMKK22}.
    }
    \label{fig:usg_ssg}
    \vspace{-2mm}
\end{figure}

\vspace{-4mm}
\paragraph{2) USG Generation across Multiple Modalities.}
To evaluate our model's performance in parsing USG across multiple modalities, we conduct experiments using a collection of pair-wise multimodal datasets. 
As a baseline, we adopt a pipeline approach, applying the best SG parser for each modality independently and then combining them together. 
As shown in Tab. \ref{tab:multimodal}, our model, trained exclusively on corresponding multimodal data, achieves superior USG generation performance compared to separate SG parsers, demonstrating its ability to leverage cross-modal complementary information to enhance accuracy.
Furthermore, joint training on all multimodal datasets consistently achieves the highest performances, underscoring USG-Par's effectiveness in generating USGs across diverse multimodal scenarios.

\vspace{-1mm}
\subsection{Ablations and Discussions}
\vspace{-1mm}
Taking one step further, here we give more discussions and in-depth analyses to reveal how the system advances.

\vspace{-4mm}
\paragraph{1) Probing Advantage of USG over Singleton SG Representations.}
Fig. \ref{fig:usg_ssg} compares the performance of multimodal tasks with and without USG integration. 
Across tasks—multimodal relation extraction (MRE), emotion detection (ED), and 3D visual QA (3DVQA)—the incorporation of USG consistently improves results over baselines. 
Additionally, USG shows superior performance compared to MSG, which is constructed through a pipeline approach, highlighting the advantages of USG representations.

\begin{table}[!t]
  \centering
\fontsize{8}{9}\selectfont
\setlength{\tabcolsep}{2.2mm}
\begin{tabular}{lccccc}
\hline
\multirow{1}{*}{\bf LN}& \multirow{1}{*}{\bf Filter
} & \multirow{1}{*}{$\mathcal{L}_{ass}$} & $\mathcal{S}-\mathcal{I}/\mathcal{V}/\mathcal{D}$  & $\mathcal{I}-\mathcal{V}$ & $\mathcal{I}-\mathcal{D}$  \\ 
\hline
$\times$ & \checkmark & \checkmark  & 4.6 / 3.8 / 1.7   & 18.2  &  12.4 \\
\checkmark & $\times$ & \checkmark  & 12.5 / 11.7/ 11.1   & 22.3  & 18.2    \\
\checkmark & \checkmark & $\times$  & 10.7 / 10.8 / 11.4   & 22.8  &  19.1   \\
\cdashline{1-6}
\checkmark & \checkmark & \checkmark  & 13.6 / 13.9 / 12.0   & 24.3 &  20.7    \\
\hline
\end{tabular}
\vspace{-2mm}
\caption{
Ablation study of object associator. ``LN'' means the linear transformation. 
Association accuracy@5 scores are reported.
}
\vspace{-5mm}
\label{tab:ass_design}
\end{table}

\vspace{-4mm}
\paragraph{2) The Necessity of Each Component of Object Associator.}
Tab. \ref{tab:ass_design} presents an ablation study on each component of the object associator.
Firstly, removing the transformation linear layer—computing cosine similarity directly without modality-specific transformations—leads to the lowest performance, as the model struggles to effectively associate objects across modalities.
Incorporating the filter further enhances performance by excluding low-confidence pairs.
Finally, adding a supervised signal to the association matrix significantly improves guidance for both the filter learner and the linear layer, enabling more precise performance.

\vspace{-4mm}
\paragraph{3) The Necessity of Each Component of RPC.}
Tab. \ref{tab:rpc} presents ablation studies on the effectiveness of each component of RPC. 
Our findings indicate all three components contribute to the performance. 
Notably, removing the projection and RAC layers leads to model divergence, resulting in no correct predictions, as object queries lack essential pairwise information and contain only category details. Additionally, removing the pair loss, which encodes critical information on pair distributions to support pair proposal matrix learning, also degrades performance.

\vspace{-5mm}
\paragraph{4) The Architecture of Relation Decoder.}
We evaluate different architectures for the relation decoder.
The results are shown in Tab. \ref{tab:rel_design}. 
We find that the transformer-based architecture outperforms the MLP-based approach. 
Additionally, integrating pairwise information with contextualized input through cross-attention further improves performance by preserving more input details.

\begin{table}[!t]
  \centering
\fontsize{8}{9}\selectfont
\setlength{\tabcolsep}{1.0mm}
\begin{tabular}{lcccccc}
\hline
\multirow{2}{*}{\bf Proj.} & \multirow{2}{*}{\bf RAC} & \multirow{2}{*}{\bf $\mathcal{L}_{pair}$} & \bf PSG & \bf PVSG  & \bf 3DDSG & \bf FACTUAL \\
\cmidrule(r){4-4} \cmidrule(r){5-5} \cmidrule(r){6-6} \cmidrule(r){7-7}
& & & \bf R/mR@50 & \bf R/mR@50 & \bf R/mR@50 &\bf  Set Match \\ 
\hline
\checkmark & \checkmark & $\times$ & 38.4 / 31.4  & 4.4 / 1.8 &  18.9 / 14.0  & 79.1  \\
\checkmark & $\times$ & \checkmark & 32.5 / 26.5  & 2.3 / 1.1&  17.9 / 7.2 & 76.3  \\
$\times$ & \checkmark & \checkmark & 36.5 / 28.5  & 4.2 / 1.5 &  18.6 / 14.9  & 78.9 \\
$\times$ & $\times$ & \checkmark & 2.5 / 1.9  & 0.5 / 0.4 &  1.6 / 1.2  & 64.5  \\
\cdashline{1-7}
\checkmark & \checkmark & \checkmark & 44.6 / 40.9  & 5.4 / 2.3 & 21.8 / 15.4  & 80.4 \\

\hline
\end{tabular}
\vspace{-2mm}
\caption{
Ablation study of the RPC.
``Proj.'' denotes the subject/object projector, and ``RAC'' is the two-way relation-aware cross-attention module. 
The evaluation is exclusively performed on the corresponding dataset.
}
\vspace{-3mm}
\label{tab:rpc}
\end{table}

\begin{table}[!t]
  \centering
\fontsize{8}{9}\selectfont
\setlength{\tabcolsep}{1.8mm}
\begin{tabular}{lcccc}
\hline
\multirow{2}{*}{\bf Architecture} & \bf PSG & \bf PVSG  & \bf 3DDSG & \bf FACTUAL \\
\cmidrule(r){2-2} \cmidrule(r){3-3} \cmidrule(r){4-4} \cmidrule(r){5-5}
& \bf R/mR@50 & \bf R/mR@50 & \bf R/mR@50 & \bf Set Match \\
\hline
MLP & 34.1 / 20.7  & 3.6 / 1.0 &  12.1 / 7.3  & 61.2 \\
w/o $F_{CA}^{rel}$ & 39.6 / 28.4  & 4.5 / 1.6 &  16.7 / 10.0 & 73.5  \\
Ours & 44.6 / 40.9  & 5.4 / 2.3 & 21.8 / 15.4  & 80.4    \\
\hline
\end{tabular}
\vspace{-2mm}
\caption{
Different Architectures for relation decoder. 
``w/o $F_{CA}^{rel}$'' denotes removing the cross-attention layers. 
}
\vspace{-5mm}
\label{tab:rel_design}
\end{table}

\vspace{-5mm}
\paragraph{5) The Impact of Text-centric Scene Contrastive Learning.}
In Tab. \ref{tab:multimodal}, we compare the model equipped with and without contrastive learning.
The results indicate that applying contrastive learning yields consistent improvements in USG generation across all multimodal datasets. This suggests that contrastive learning effectively mitigates modality gaps, leading to enhanced overall performance.

\vspace{-2mm}
\subsection{Qualitative Case Study with Visualization}

\vspace{-1mm}

\begin{figure}[!t]
    \centering
    \includegraphics[width=0.99\linewidth]{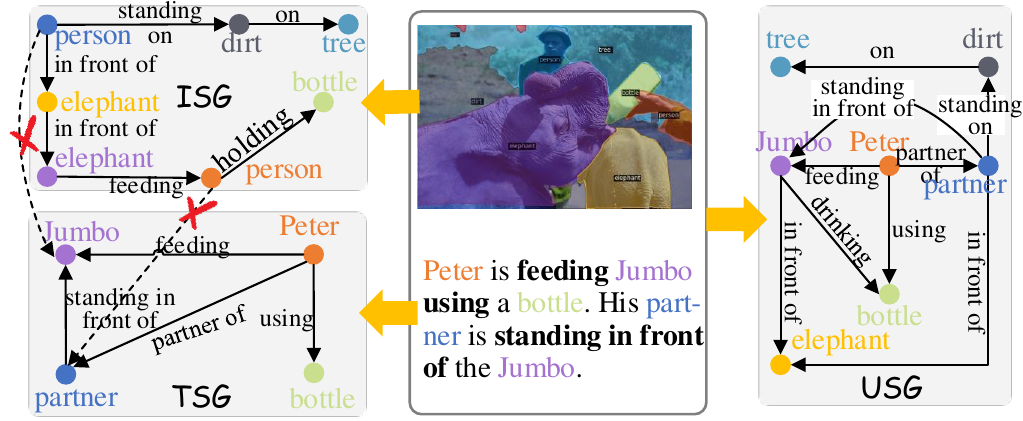}
    \vspace{-3mm}
    \caption{The USG derived from image and text and wrong association built between ISG and TSG by the pipeline method.}
    \label{fig:t_i_case}
    \vspace{-5mm}
\end{figure}

Finally, we visualize a USG of both image and text inputs from our system.
As in Fig. \ref{fig:t_i_case}, the pipeline approach, which applies separate SG parsing of each modality and then combines them, often leads to incorrect associations, such as an erroneous relation between ``person'' and ``Jumbo''.
Conversely, we find that USG can offer a more comprehensive scene representation by accurately aligning cross-modal objects and integrating information from both modalities. 
For instance, the USG correctly identifies ``Peter'' as the person holding the bottle and ``Jumbo'' as \emph{fed elephant} while also effectively integrating other visual elements, e.g., \emph{trees} and \emph{dirt}. 
we provide more visualizations in the Appendix.

\vspace{-2mm}
\section{Related Work}
\vspace{-2mm}

Over decades, SGs have garnered substantial research attention \cite{Motifs-ZellersYTC18,VG-KrishnaZGJHKCKL17,3DDSG-WaldDNT20,FACTUAL-LiCZQHLJT23}, where various definitions of SG representations under different modalities and settings are developed \cite{VG-KrishnaZGJHKCKL17,OpenImage-KuznetsovaRAUKP20,AG-JiK0N20,3DDSG-WaldDNT20,SchusterKCFM15}, including 
image SG \cite{JohnsonKSLSBL15,VG-KrishnaZGJHKCKL17,Motifs-ZellersYTC18},
textual SGs \cite{SchusterKCFM15,FACTUAL-LiCZQHLJT23}
video \cite{VG-KrishnaZGJHKCKL17},
3D \cite{3DDSG-WaldDNT20,SGFN-WuWTNT21}, 
and even more settings such as panoptic SG \cite{PSG-YangAGZZ022,PVSG-YangPLGCL0ZZLL23,yang2023d} and ego-view SG \cite{RodinF0TF24}, etc. 
SGs can accurately capture the semantics of a scene while filtering out undesired visual information. 
Thus, SGs have been widely applied to various downstream tasks \cite{abs-2403-19098,abs-2403-17846,TaharaSNI20,SchusterKCFM15}.
While almost all existing SG research is confined to modeling within a single modality, we realize that real-world scenarios necessitate a universal SG representation capable of expressing information from various modalities through a unified cross-modal perspective. 
This need is particularly pressing with the development of multimodal generalist and agent communities \cite{Wu0Q0C24,LiuLWL23a,0001Z24,abs-2408-06926}, where an increasing number of applications require the ability to understand and process multimodal information. 
Therefore, this paper for the first time explores a novel USG representation. 
Despite the existence of various SG generation methods for different SG types, there should currently not be a specialized approach for universally parsing SGs across modalities. 
Specifically, a unified model architecture is required for both modeling modality-invariant SG information and efficiently preserving the complementary modality-specific scene.

\vspace{-1mm}
\section{Conclusion}
\vspace{-1mm}

This paper presents a Universal Scene Graph (USG), a novel representation that characterizes comprehensive semantic scenes from any combination of modality inputs, encompassing both modality-invariant and modality-specific aspects.
To generate USG effectively, we develop USG-Par, a niche-targeting parser for end-to-end USG generation.
USG-Par addresses the critical challenges of cross-modal object alignment and out-of-domain generalization by incorporating an object associator that bridges modality gaps and a text-centric scene contrasting learning mechanism that mitigates domain imbalances.
Through extensive experiments, we demonstrate that USG provides a more powerful and comprehensive semantic scene representation compared to standalone SGs.
Also USG-Par achieves superior efficacy, offering a strong benchmark method for USG.

\newpage

{
    \small
    \bibliographystyle{ieeenat_fullname}
    \bibliography{main}
}






\clearpage
\setcounter{page}{1}
\maketitlesupplementary
\appendix

\section*{Overview}

The appendix presents more details and additional results not included in the main paper due to page limitation. The list of items included are:

\begin{compactitem}
    \item Specification on Task Definition and Setups in $\S$\ref{app:problem_setting};
    \item Full-version Related Work in $\S$\ref{app:full-related-work};
    \item Limitations and Future Direction in $\S$\ref{app:limitation-future};
    \item Extended Framework Details in $\S$\ref{app:framrework};
    \item Detailed Experimental Settings in $\S$\ref{app:settings};
    \item Extended Experimental Results in $\S$\ref{app:experiments}.
\end{compactitem}

\section{Specification on Task Definition and Setups}
\label{app:problem_setting}

\subsection{SG Structure}

Here, we provide a detailed description of the nodes and edges in the USG.
The USG is formally represented as $\mathcal{G}^{\mathcal{U}} = \{\mathcal{O}, \mathcal{R}\}$, where $\mathcal{O} = \{\mathcal{O}^{*}\}$, $* \in \{\mathcal{I}, \mathcal{V}, \mathcal{D}, \mathcal{S} \}$ represents the set of objects across all modalities. 
Each node involves a category label $c_i^o \in \mathbb{C}^{\mathcal{O}}$ and a segmentation mask $m_i$. 
For instance, as illustrated in Fig. \ref{fig:T-I-USG}, the objects node set $\mathcal{O}$ in the USG comprises of textual objects node set $\mathcal{O}^{\mathcal{S}}$ in the TSG and visual objects node set $\mathcal{O}^{\mathcal{I}}$ in the ISG. 
$\mathcal{R} = \{\mathcal{R}^{*}, \mathcal{R}^{* \times \diamond}\}$, $*, \diamond \in \{\mathcal{I}, \mathcal{V}, \mathcal{D}, \mathcal{S} \} $ \text{and} $ * \ne \diamond $. $\mathcal{R}^{*}$ includes both intra-modality relationships and inter-modality associations $\mathcal{R}^{* \times \diamond}$.
We define the existence of inter-modality associations between objects from different modalities if they correspond to the same underlying object described in distinct modalities.
For example, as shown in Fig. \ref{fig:T-I-USG}, the textual object ``\textit{Peter}'' in the TSG should correspond to the visual object ``\textit{person}'' in the ISG. 
Similarly, as depicted in Fig. \ref{fig:I-3D-USG},  the ``\textit{sofa}'' in the 3DSG aligns with the ``\textit{sofa}'' in the ISG.
When inter-modality associations exist, the corresponding objects are merged into a unified node, as shown in Fig.~\ref{fig:T-I-USG}, with the example of the ``\textit{headphones}''
This merged node represents the object across multiple modalities, retaining a single category label. 
Typically, the object name from the textual modality is prioritized for its flexibility and precision in description. 
Similarly, the relation predicate is preferentially adopted from the TSG, as it often provides a more descriptive and accurate representation. 
For instance, in Fig.~\ref{fig:T-I-USG}, the relationship between ``\textit{Peter}'' and ``\textit{sofa}'' in the USG is ``\textit{relax on}'' derived from the TSG, rather than ``\textit{lying}'' which might be less descriptive.
Despite merging nodes, the segmentation masks from all modalities are preserved. This ensures that each modality's unique contribution to the object's representation is maintained within the USG.

In addition, to parse the USG for scenes derived from video and other modalities, we first establish association relations between nodes from other modalities and the objects in each frame of the VSG. 
For instance, as illustrated in Fig.~\ref{fig:T-V-USG}, the objects ``\textit{Peter}'', ``\textit{sofa}'' and ``\textit{iPhone}'' from the TSG are associated with the objects in every frame of the VSG.
To ensure the USG comprehensively represents the scene described by the video and other modalities, the scene from the other modalities is added as the first frame in the USG. 
The remaining frames correspond to the frame-level scene graph representations from the VSG. 
This paradigm advances in integrating multimodal information much more seamlessly, enriching the holistic representation of the scene within the USG framework.

\subsection{Combination of All Possible Modalities}

\begin{figure}
    \centering
    \includegraphics[width=0.99\linewidth]{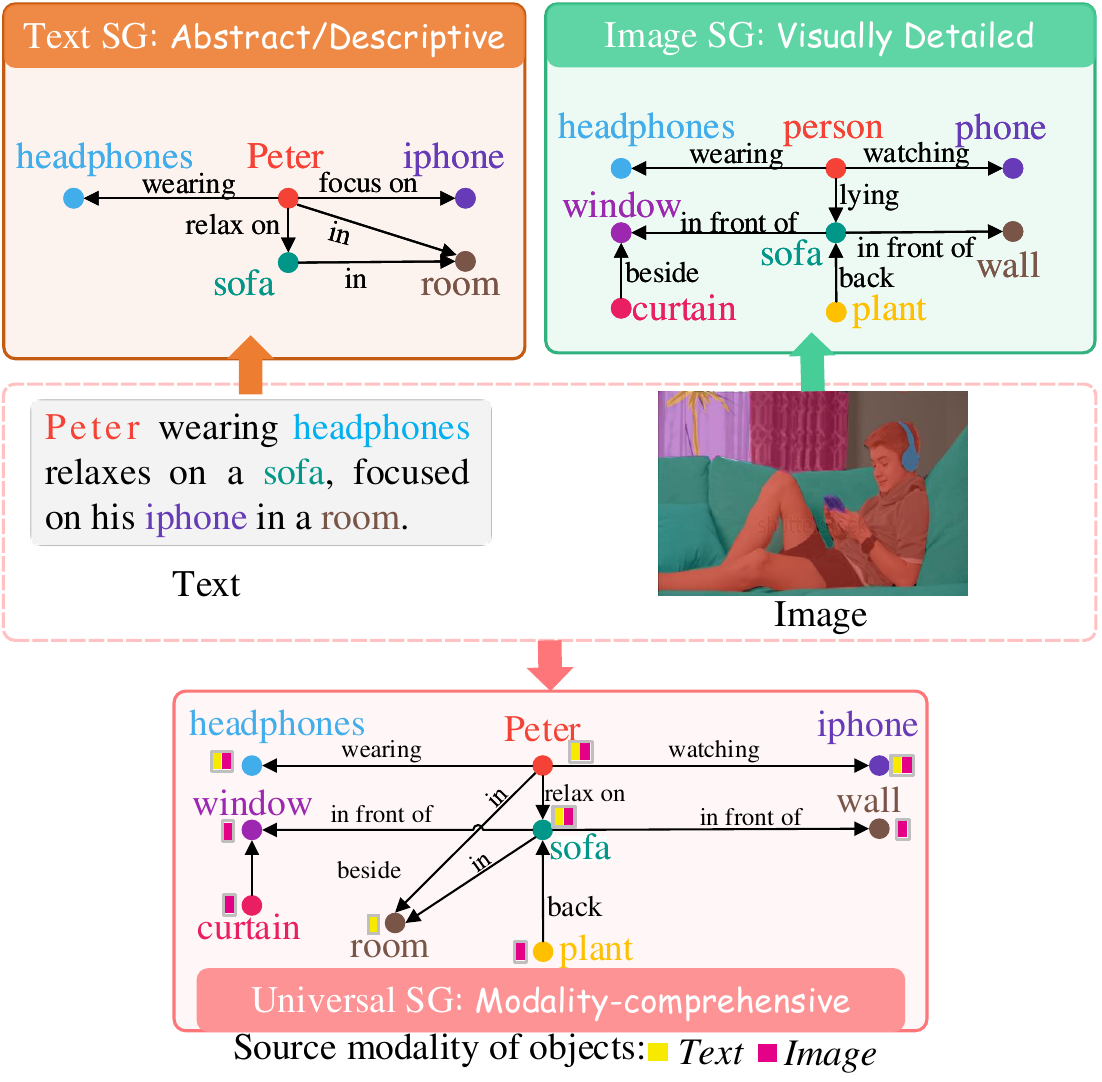}
    \vspace{-2mm}
    \caption{Illustration of USG generated from text and image scenes.}
    \label{fig:T-I-USG}
    \vspace{-4mm}
\end{figure}

\begin{figure}
    \centering
    \includegraphics[width=0.99\linewidth]{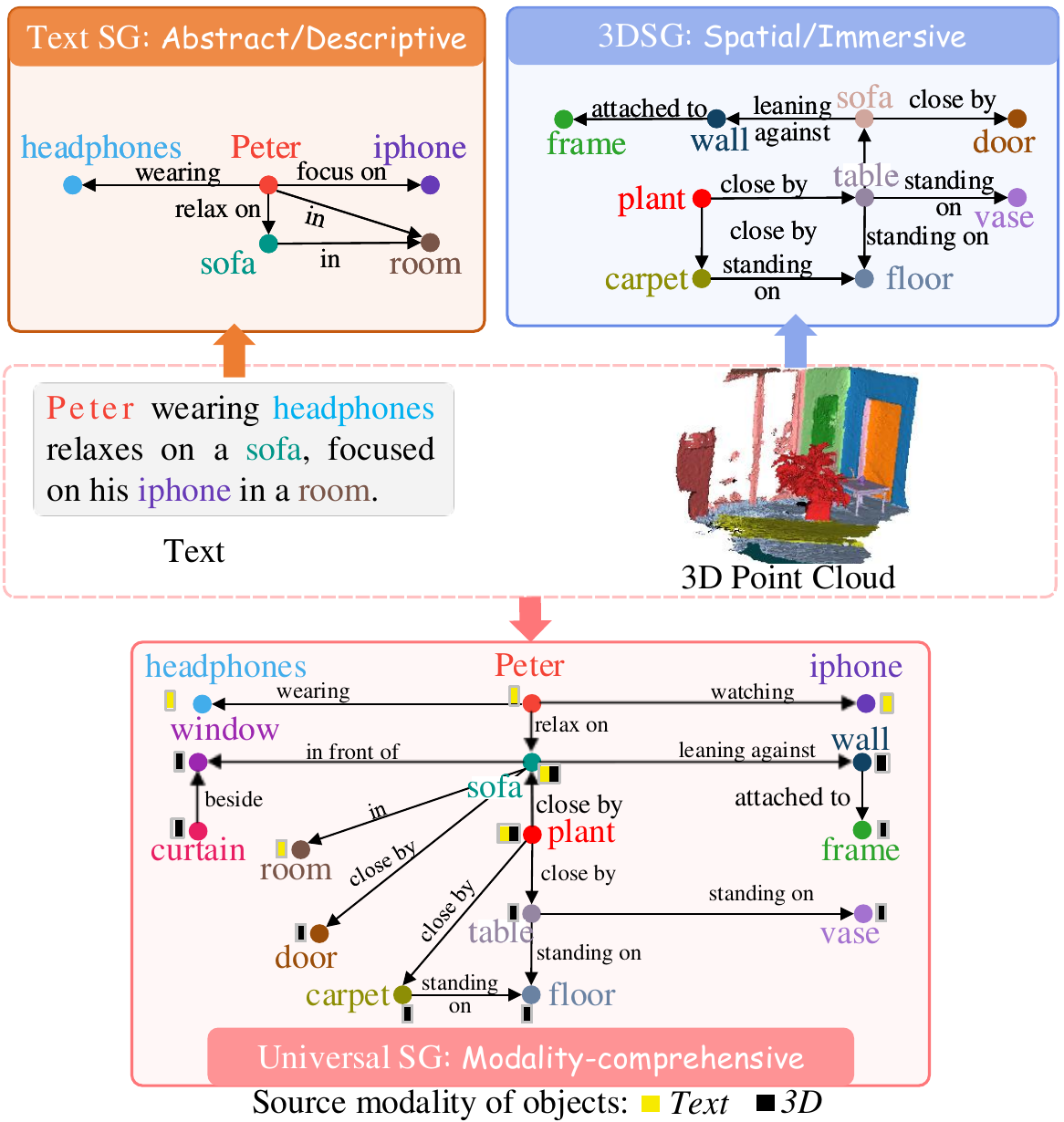}
    \vspace{-2mm}
    \caption{Illustration of USG generated from text and 3D scenes.}
    \label{fig:T-3D-USG}
    \vspace{-2mm}
\end{figure}

\begin{figure}
    \centering
    \includegraphics[width=0.99\linewidth]{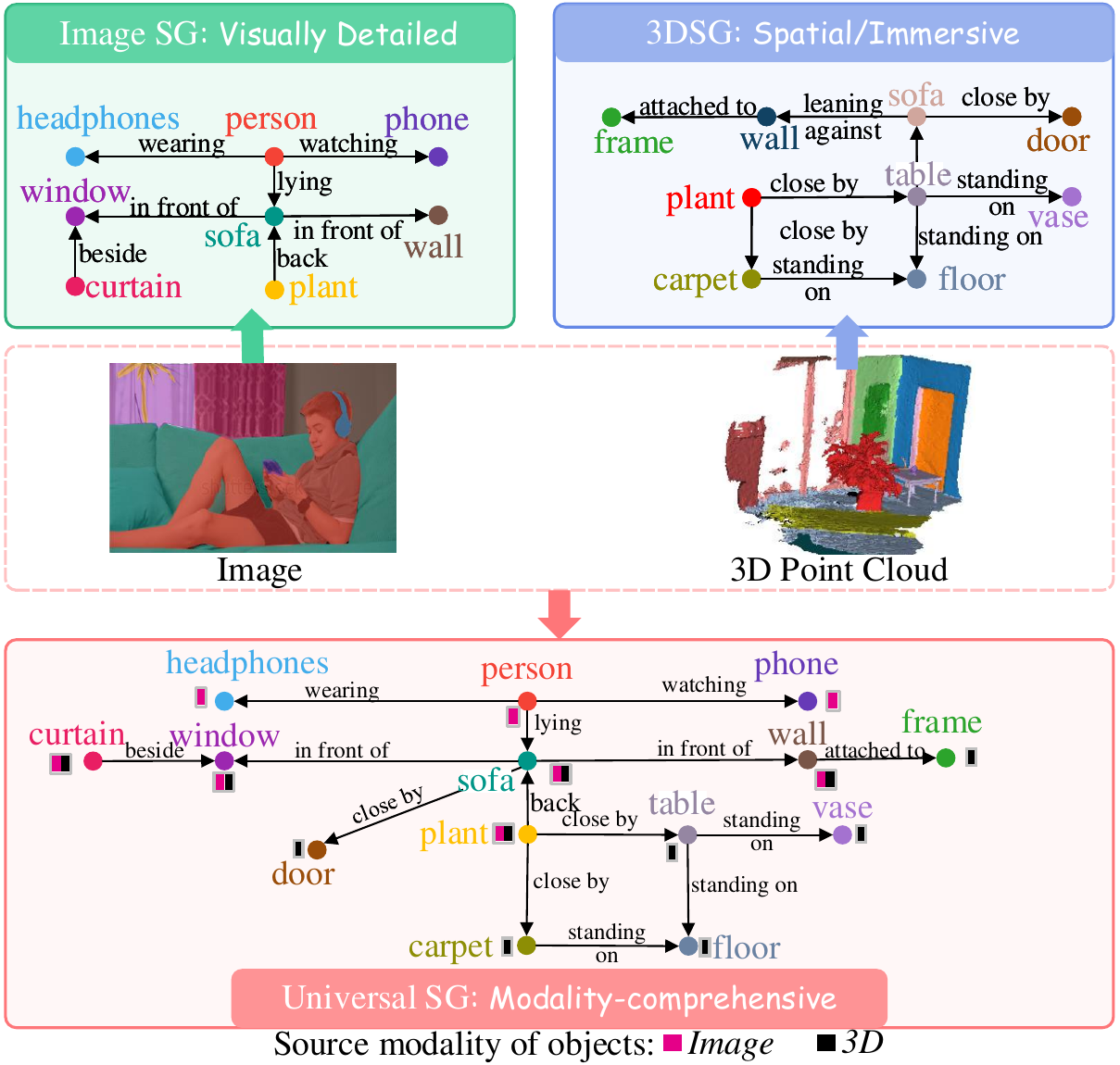}
    \vspace{-2mm}
    \caption{Illustration of USG generated from image and 3D scenes.}
    \label{fig:I-3D-USG}
    \vspace{-2mm}
\end{figure}

\begin{figure*}
    \centering
    \includegraphics[width=0.75\linewidth]{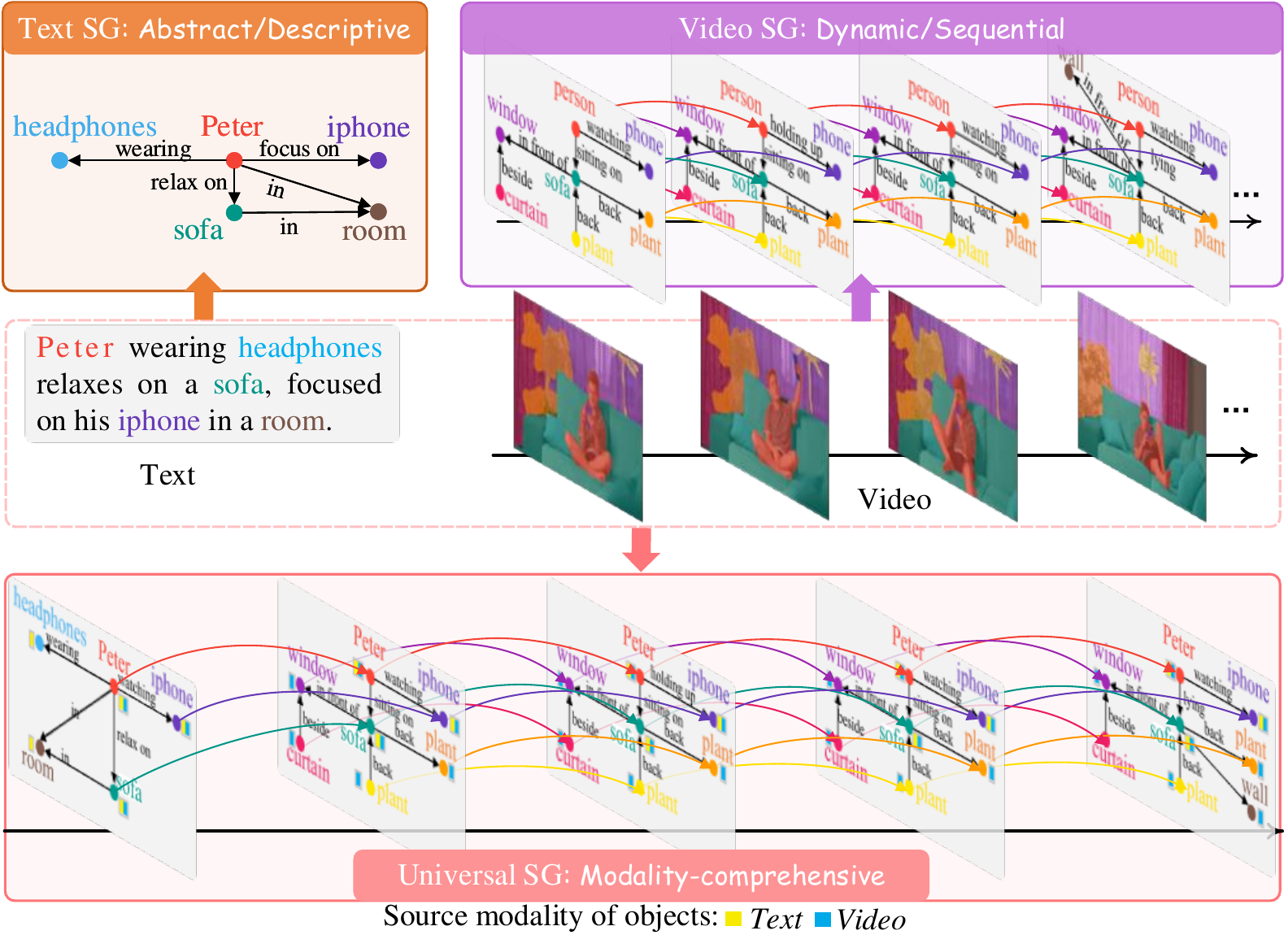}
    \vspace{-2mm}
    \caption{Illustration of USG generated from text and video scenes.}
    \label{fig:T-V-USG}
\end{figure*}

\begin{figure*}
    \centering
    \includegraphics[width=0.75\linewidth]{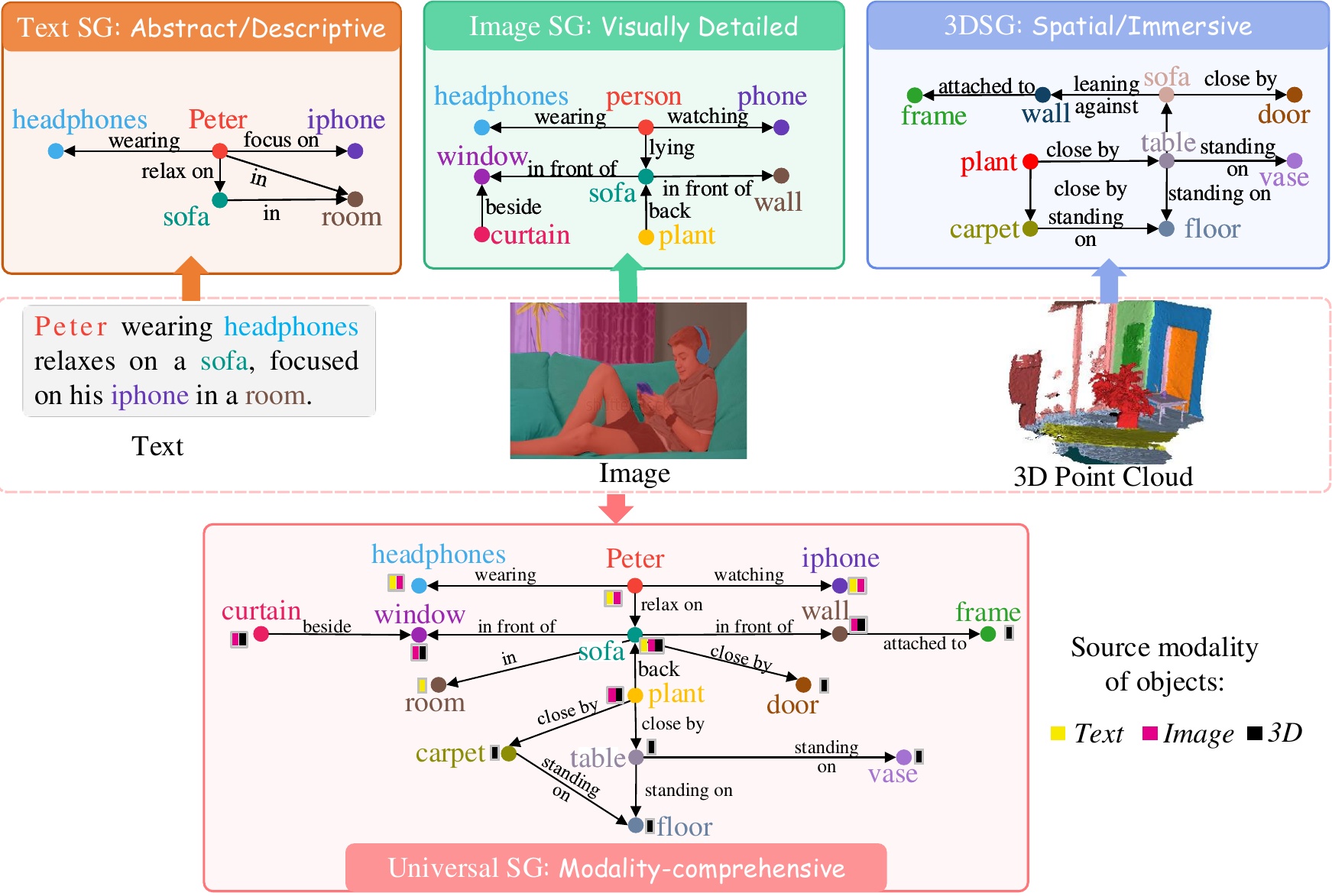}
    \vspace{-2mm}
    \caption{Illustration of USG generated from text, image and 3D scenes.}
    \label{fig:T-I-3D-USG}
    \vspace{-4mm}
\end{figure*}

\begin{figure*}[!t]
    \centering
    \includegraphics[width=0.99\linewidth]{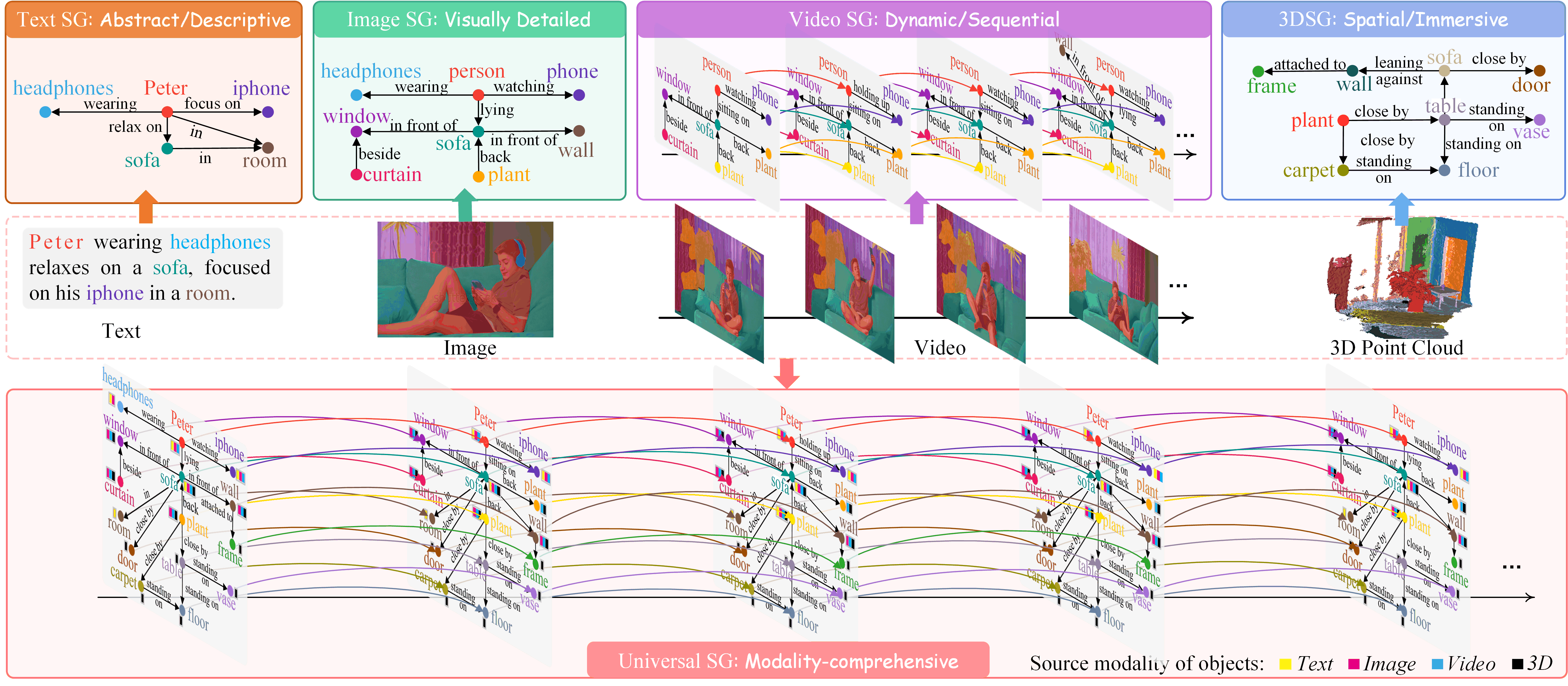}
    \vspace{-2mm}
    \caption{Illustration of USG generated from the text, image, video, and 3D in a stereoscopic viewpoint.
    This is also the full version of the first illustration shown in the Introduction section of the main article.
    Best viewed via zooming in.
    }
    \label{fig:enter-label}
    \vspace{-2mm}
\end{figure*}

Here we provide illustrations of the USG obtained under different modal combinations:
\begin{compactitem}
    \item \textbf{Text-Image}: in Fig.\ref{fig:T-I-USG}.
    \item \textbf{Text-3D}: in  Fig.\ref{fig:T-3D-USG}.
    \item \textbf{Text-Video}: in Fig.\ref{fig:T-V-USG}.
    \item \textbf{Image-3D}: in Fig.\ref{fig:I-3D-USG}.
    \item \textbf{Text-Image-3D}: in Fig.\ref{fig:T-I-3D-USG}.
    \item \textbf{Text-Image-Video-3D (Complete combination)}: in Fig.\ref{fig:enter-label} we provide a full illustration of USG obtained from the total 4 modalities, i.e., text, image, video, and 3D. 
\end{compactitem}

\section{Full-version Related Work}
\label{app:full-related-work}

\subsection{SG Representation and Definition}

Research on SG generation \cite{Motifs-ZellersYTC18,VG-KrishnaZGJHKCKL17,3DDSG-WaldDNT20,FACTUAL-LiCZQHLJT23} has long been a significant focus within the relevant community, aiming to deeply understand environmental scenes by not only recognizing individual objects but also the semantic relationships between them. 
Over decades of development, SGs have garnered substantial research attention and efforts, 
where various definitions of SG representations under different modalities and settings are developed \cite{VG-KrishnaZGJHKCKL17,OpenImage-KuznetsovaRAUKP20,AG-JiK0N20,3DDSG-WaldDNT20,SchusterKCFM15}.
Initially, centered on the static vision, researchers pioneered image SG \cite{JohnsonKSLSBL15,VG-KrishnaZGJHKCKL17,Motifs-ZellersYTC18}, where nodes represent objects and edges denote the relationships between them. 
Subsequently, textual SGs \cite{SchusterKCFM15,FACTUAL-LiCZQHLJT23} are proposed to acknowledge that the textual modality can also convey a complete scene. 
Later studies extended SG representations to other data modalities, including video \cite{VG-KrishnaZGJHKCKL17} and 3D \cite{3DDSG-WaldDNT20,SGFN-WuWTNT21}, and even to more settings such as panoptic SG \cite{PSG-YangAGZZ022,PVSG-YangPLGCL0ZZLL23} and ego-view SG \cite{RodinF0TF24}, etc. 
SGs can accurately capture the semantics of a scene while filtering out undesired visual information. 
Moreover, different modalities possess distinct characteristics, allowing SGs to model semantic scenes with subtly different traits. 
Thus, SGs have been widely applied to various downstream tasks \cite{abs-2403-19098,abs-2403-17846,TaharaSNI20,SchusterKCFM15}.

As previously emphasized, the definitions of SGs across differing modalities result in varied features and strengths. 
While almost all existing SG research is confined to modeling within a single modality, we recognize that real-world scenarios necessitate a universal SG representation capable of expressing information from various modalities through a unified cross-modal perspective. 
This need is particularly pressing with the development of multimodal generalist and agent communities \cite{Wu0Q0C24,LiuLWL23a,0001Z24,abs-2408-06926}, where an increasing number of applications require the ability to understand and process multimodal information. 
Therefore, this paper explores a novel USG representation for the first time.

\subsection{SG Generation Methods}

Historically, SG generation methods can be broadly categorized into two main groups: two-stage methodsand one-stage methods.
The two-stage methods \cite{LiCXYW023,LiZW021,GPS-Net-LinDZT20,LuRCKY0TV21,SudhakaranDK023,XuZCF17,Motifs-ZellersYTC18,ZhengL0ZSG22}  involve training separate object detection and relation prediction models sequentially. 
Typically, these methods rely on off-the-shelf object detectors, such as Faster R-CNN \cite{Faster-RCNN-RenHGS15}, to detect N object queries. 
Subsequently, features such as appearance, spatial information, labels, depth, and masks are extracted for all possible combinations of detected objects. 
These features are then fed into the relation prediction model to infer relationships between each object pair. 
Despite achieving high relation extraction performance, the inherent limitations of the pipeline approaches, particularly the separate training of components, lead to significant model complexity.

To address this issue, recent research has shifted towards one-stage methods \cite{OED-WangLCL24,Pair-Net-10634834,RelTR-CongYR23,RepSGG-10531674}, where the object detector and relation extractor are trained in an end-to-end manner. 
Early studies proposed fully convolutional SG generation models \cite{LiuYMB21} and adopted pixel-based approaches \cite{NewellD17}. 
Following the success of DETR \cite{DETR-CarionMSUKZ20}, a Transformer-based one-stage object detector, many one-stage SGG studies \cite{DSG-DETR-FengMNMT23,EGTR-ImNPLP24} have adopted similar approaches. 
These methods effectively model SG generation by introducing object queries or triplet queries. 
For instance, RelTR \cite{RelTR-CongYR23} introduced paired subject and object queries, while SGTR \cite{SGTR-LiZ022} proposed compositional queries decoupled into subjects, objects, and predicates. 
PairNet \cite{Pair-Net-10634834} designed separate relation and object queries, and PSGTR \cite{PSG-YangAGZZ022} directly introduced triplet queries to detect triplets without relying on an object detector.

Beyond architectural designs, several studies \cite{TEMPURA-NagMTR23,Teng0LW21,OED-WangLCL24,APT-Arnab0S21} have focused on leveraging modality-specific characteristics to enhance model performance. 
In the context of VSG generation, modeling spatio-temporal features has garnered significant attention. 
For example, TRACE \cite{Teng0LW21} employs a hierarchical tree structure to aggregate spatial context, and \citep{Arnab0S21} utilizes message passing in a spatio-temporal graph to enhance feature representation.
For 3DSG generation, some researches \cite{feng2023exploring,FengH0WGM23} focus on leveraging the spatial layout clues to enhance the 3DSG generation performance.

Additionally, to improve performance, many works are no longer limited to using only visual appearance.
External knowledge has been incorporated to further improve SG generation performance \cite{YuLMD17,CogTree-YuCW0W21,feng2023exploring,FengH0WGM23}. 
This includes statistical priors \cite{YuLMD17}, such as co-occurrence frequencies, and commonsense knowledge \cite{abs-2311-12889,ChenRL23,HiKER-SGG-0009SCSX24} extracted from sources like Wikipedia or ConceptNet \cite{ConceptNet-SpeerCH17}.

However, existing SG generation methods remain modality-specific, with no approach capable of supporting SG generation across different modalities. 
This limitation highlights the emergency of developing a universal SG generation method.

\vspace{-1mm}
\section{Limitations and Future Direction}
\label{app:limitation-future}

\vspace{-1mm}
\subsection{Potential Limitations}
\label{app:limitations}
\vspace{-1mm}

Despite its contributions, this work has several limitations:
Firstly, the proposed method faces challenges in associating objects across different modalities in highly complex and densely populated scenes. 
For instance, distinguishing between multiple similar individuals or matching objects with their textual object names often requires external commonsense knowledge, which is beyond the current scope of the model.
Secondly, in video scenes, the method struggles with particular long-term understanding, particularly in object tracking and relation recognition over extended temporal sequences. While our current dataset does not include particularly long videos, such scenarios are common in real-world applications. Addressing this limitation presents a valuable direction for future research.

\vspace{-1mm}
\subsection{Future Work on USG}
\vspace{-1mm}

Going forward on the USG we introduced in this work, we believe the following aspects should be worth exploring.

First, the USG has significant potential for enhancing the capabilities of multimodal large language models (MLLMs). 
As a modality-invariant universal representation, the USG facilitates fine-grained alignment across different modalities, including object-level and relation-level correspondences.
Inspired by the concept of knowledge masking \cite{SunWLFTWW20}, which focuses on learning more structured knowledge by masking phrases and named entities rather than individual sub-words, USG can inject fine-grained, structured semantic knowledge across modalities into MLLMs. 
This approach enables the alignment of semantic information at a granular level, fostering a deeper and more precise understanding of cross-modal content.
Specifically, pre-training tasks can be designed by masking and predicting various types of nodes in the USG, parsed from multimodal inputs. These nodes may correspond to objects, relationships, or attributes, and their structured representation allows the model to learn modality-invariant features effectively.
This strategy not only improves the alignment across modalities but also strengthens the model’s reasoning and generalization capabilities.

Beyond MLLMs, the USG can also serve as a foundation for numerous downstream applications. 
In robotics, USG could facilitate embodied AI tasks, such as planning \cite{ConceptGraphs-GuKMJSARPECGMTT24} and navigation \cite{abs-2403-17846,abs-2304-03696}, by providing a universally structured understanding of dynamic, multimodal environments. 
Moreover, in creative applications like content generation, USG could bridge visual and textual modalities to produce contextually coherent and semantically rich outputs, such as image-to-text descriptions or cross-modal story generation.

Looking ahead, the universality of USG should lead to unified multimodal benchmarks, where diverse tasks can be evaluated under a consistent framework. 
This would drive innovation in creating truly generalizable AI systems capable of reasoning across multiple domains and modalities. 
Developing such systems could redefine the boundaries of multimodal AI, enabling applications that require deep contextual understanding, such as virtual reality simulations, autonomous systems, and interactive AI agents.

\section{Extended Framework Details}
\label{app:framrework}

In this part, we try to give a more comprehensive picture of our USG generation framework, as an extension to the description in the main article.

\begin{figure}[!th]
    \centering
    \includegraphics[width=0.98\linewidth]{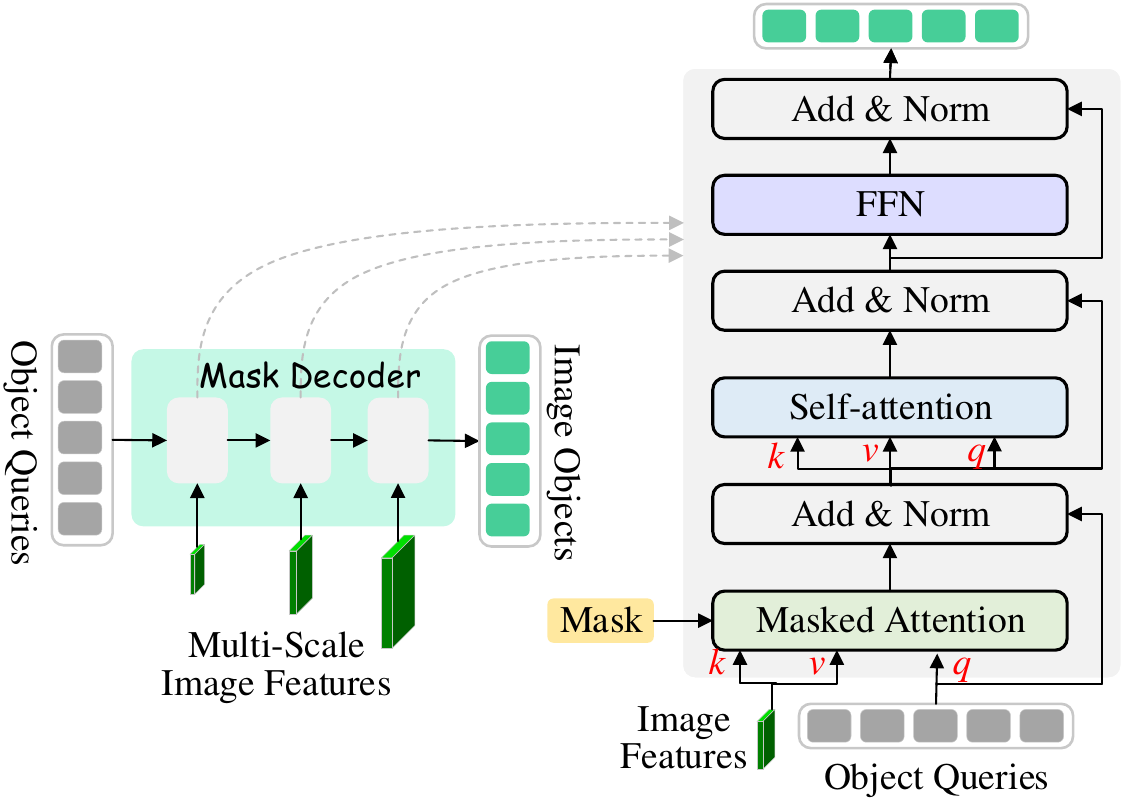}
    \vspace{-2mm}
    \caption{The framework of the mask decoder: multi-scale image features are integrated to refine image object queries, following a similar approach for other modalities such as text, video, and 3D.}
    \label{fig:mask-decoder-img}
    \vspace{-2mm}
\end{figure}

\subsection{Mask Encoder}
As depicted in Fig. \ref{fig:mask-decoder-img}, the randomly initialized object queries are fed into the mask decoder, in which the multi-scale image features are also injected by masked cross-attention to help refine the object query representations. 
Specifically, following \cite{mask2former-ChengMSKG22}, we perform masked cross-attention between modality-specific features $\bm{H}^{*}$ and the corresponding object query features $\bm{X}^{*}_l \in \mathbb{R}^{N_q^{*} \times d}, * \in \{\mathcal{I},\mathcal{V},\mathcal{D},\mathcal{S}\}$ as follows:
\setlength\abovedisplayskip{3pt}
\setlength\belowdisplayskip{3pt}
\begin{equation}
    \bm{X}^{*}_l = \text{softmax}(\bm{M}^{*}_{l-1} + \bm{Q}_{l-1}^{*}{\bm{K}_{l-1}^{*\top}}) \bm{V}_{l-1}^{*} + \bm{X}_{l-1}^{*},
\end{equation}
where $N_q^{*}$ is the number of queries and $l$ is the layer index.
$\bm{X}_0^{*}$ denotes input object query features to the mask decoder.
$\bm{Q}_{l-1}^{*} = F_q(\bm{X}_{l-1}^{*})$, while $\bm{K}_{l-1}^{*} = F_k(\bm{H}^{*}) $ and $\bm{V}_{l-1}^{*} = F_v(\bm{H}^{*})$.
Here, $F_q(\cdot)$, $F_k(\cdot)$ and
$F_v(\cdot)$ are linear transformations as typically applied in attention mechanisms.
$\bm{M}^{*}_{l-1}$ is the binarized output of the resized mask prediction from the previous $l-1$-th Transformer decoder layer:
\begin{equation}
    \bm{M}_{l-1}^{*}(x,y) = \left\{\begin{array}{cc}
        0 & \text{if} \; \bm{M}_{l-1}^{*}(x,y)=1 \\
        -\infty & \text{otherwise}
    \end{array} \right.
\end{equation}
Moreover, in practice, for image, video, and 3D data, $\bm{H}^{*}$ is sampled from the multi-scale feature output $\{\bm{H}^{\mathcal{I}/\mathcal{V}/\mathcal{D}}\}_{i=1}^{3}$, while for text, we employ $\bm{H}^{\mathcal{S}}$ across different scales.
In addition, for video data, to effectively capture the temporal information across frames, we incorporate a transformer-based temporal encoder $F_{temp}$ to model the temporal relationships between objects.
After $L^{mask}$ layers, we obtain the refined object queries $\bm{Q}^{*} = \{\bm{q}_i^{*}\}_{i=1}^{N_q^*}$.

\begin{figure}[ht]
    \centering
    \includegraphics[width=0.90\linewidth]{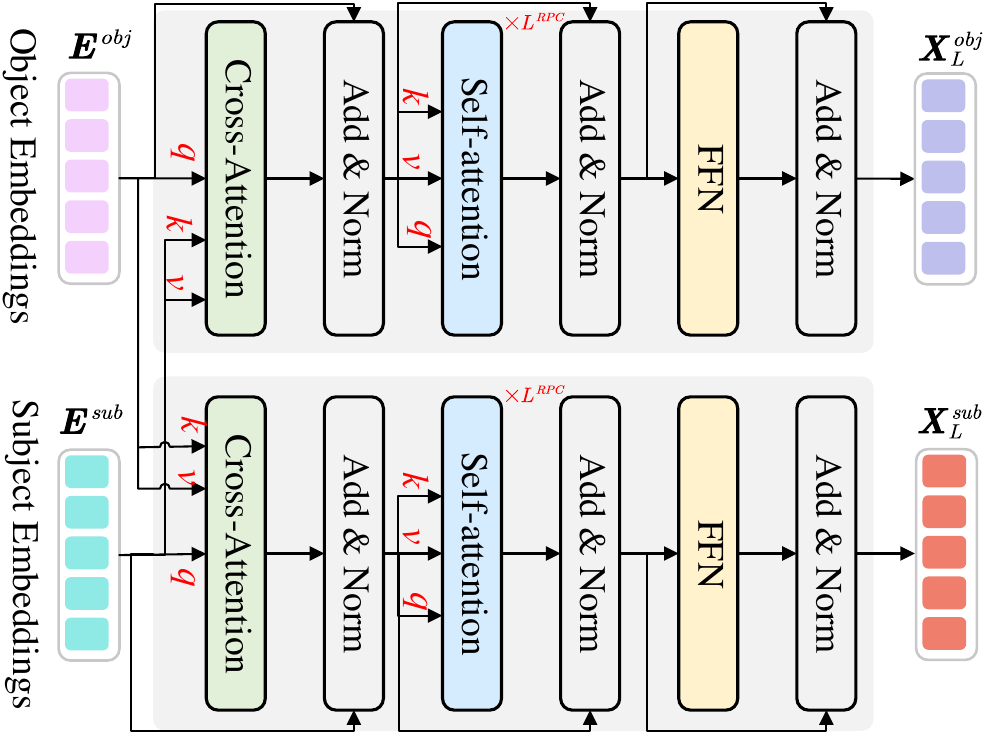}
    \vspace{-2mm}
    \caption{The framework of two-way relation-aware object/subject interaction module.}
    \label{fig:two-way-relation-aware}
    \vspace{-2mm}
\end{figure}

\subsection{Two-way Relation-aware Object/Subject Interaction}

The detailed framework of the two-way relation-aware object/subject interaction module is demonstrated in Fig. \ref{fig:two-way-relation-aware}.
The two inputs are object embeddings and subject embeddings, and then the $L^{RPC}$ layers transformer layers with cross-attention and self-attention mechanisms perform to iteratively refine subject and object features as follows:
\begin{equation}
\begin{aligned}
   \bm{X}^{sub}_{l} &= F_{\text{CA}}^{obj \rightarrow sub}(\bm{X}^{sub}_{l-1},\bm{X}^{ obj}_{l-1}, \bm{X}^{obj}_{l-1}), \\
   \bm{X}^{obj}_{l} &= F_{\text{CA}}^{sub \rightarrow obj}(\bm{X}^{ obj}_{l-1}, \bm{X}^{sub}_{l-1}, \bm{X}^{sub}_{l-1}), \\
\end{aligned}
\end{equation}
where $l$ denotes the layer index, and $\bm{X}^{sub}_{0} = \bm{E}^{sub}, \bm{X}^{obj}_{0} = \bm{E}^{obj} $. 
We define the $F_{\text{CA}}(\bm{X}, \bm{Y})$ as:
\begin{equation}
    F_{\text{CA}}(\bm{X}, \bm{Y}) = \text{softmax}(F_{q}(\bm{X})^{\top} \cdot F_{k}(\bm{Y})) \cdot F_{v}(\bm{Y}),
\end{equation}
where $F_q(\cdot)$, $F_k(\cdot)$ and
$F_v(\cdot)$ are linear transformations as typically applied in attention mechanisms.

\begin{figure}[ht]
    \centering
    \includegraphics[width=0.90\linewidth]{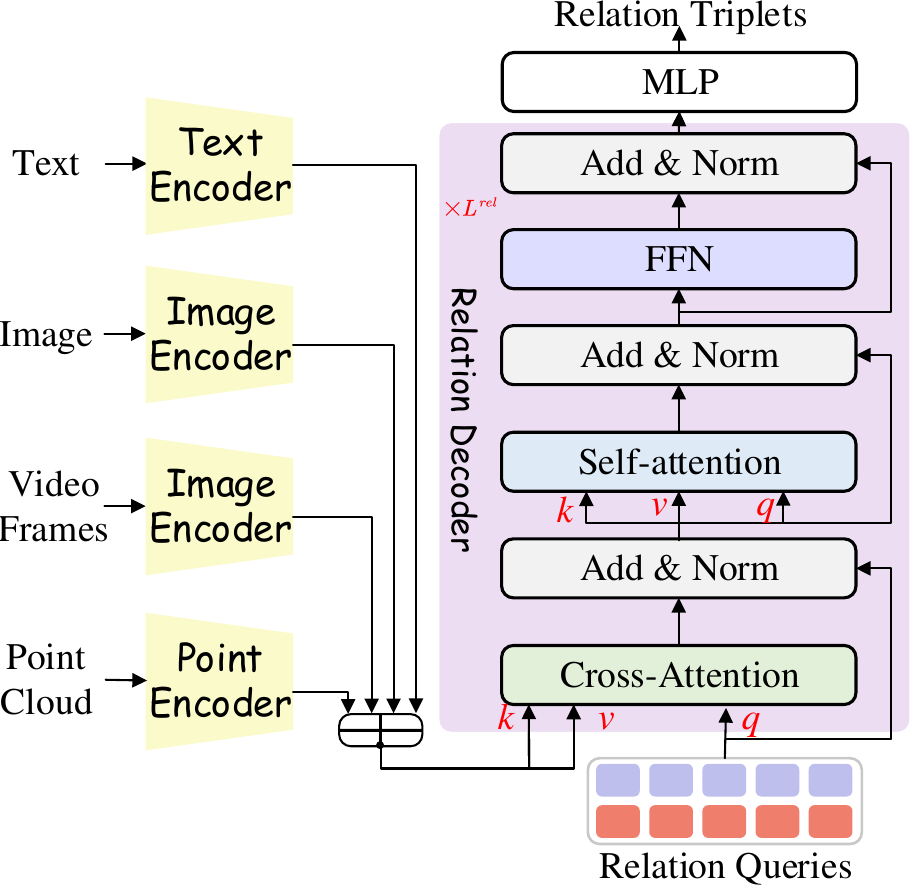}
    \vspace{-2mm}
    \caption{Illustration of relation decoder.}
    \label{fig:relation-decoder}
    \vspace{-2mm}
\end{figure}

\subsection{Relation Decoder}

As illustrated in Fig. \ref{fig:relation-decoder}, the relation encoder processes the relation queries $\bm{Q}^{rel}$ alongside contextualized information. 
Specifically, the initial relation queries are constructed by concatenating the embeddings of selected subject-object pairs.
Then, to leverage complementary contextual information from multiple modalities, we propose to fuse the multimodal features to enhance the relation extraction performance:
\begin{equation}
    \bm{H} = [\bm{H}^{\mathcal{S}};\bar{\bm{H}}^{\mathcal{I}};\bar{\bm{H}}^{\mathcal{V}};\bar{\bm{H}}^{\mathcal{D}}],
\end{equation}
where $\bm{H}$ represents the fused features, dependent on the input modalities. For example, given text, image, and 3D inputs, $ \bm{H} = [\bm{H}^{\mathcal{S}};\bar{\bm{H}}^{\mathcal{I}};\bar{\bm{H}}^{\mathcal{D}}]$.
Then, the fused features are integrated using a cross-attention mechanism to retain critical relational information:
\begin{equation}
\begin{aligned}
     \bm{X}^{rel}_{l} &= \text{F}_{CA}^{rel}(\bm{X}^{rel}_{l-1}, \bm{H}, \bm{H}) \\
     & = \text{softmax}(F_{q}(\bm{X}^{rel}_{l-1})^{\top} \cdot F_{k}(\bm{H})) \cdot F_{v}(\bm{H}), \\
\end{aligned}
\end{equation}
where $\bm{X}^{rel}_{0} = \bm{Q}^{rel}$ is the initialized relationship query features into the relation decoder.

\subsection{Inference}
\vspace{-1mm}
During inference, our framework, developed as a USG parser, supports both single-modality and multimodal input for USG generation.
For single-modality USG generation, we first perform object detection, select the most confidential relation proposals, and finally perform the relationship classification. 
For multimodal USG generation, we introduce an object associator to establish associations between object pairs across different modalities before object detection and relation classification. 
We leverage Hungarian Assignment to find the associated pairs. 
Beyond this step, the remaining procedure closely follows that of single-modality SG generation.

For open-vocabulary USG generation, we compute the cosine similarity between each predicted object query embedding and a set of class label embeddings derived from CLIP \cite{Radford2021LearningTV}. 
The final label for each object is then assigned based on the highest cosine similarity score. 
Similarly, predicate classes are determined by selecting the label with the closest cosine similarity to the text embeddings of all predicate candidates.

\section{Detailed Experimental Settings}
\label{app:settings}

\subsection{Datasets}
\label{app:datasets}

To evaluate the efficacy of USG-Par, which supports both single-modality and multi-modality scene parsing, we utilize existing single-modality datasets and a manually constructed multimodal dataset.

\subsubsection{Single-modal Dataset}
The single-modality datasets used in our experiments are categorized into the following four groups based on modality:
\vspace{-1mm}
\paragraph{Image:} 

\textbf{1) Visual Genome (VG)} \cite{VG-KrishnaZGJHKCKL17}. We follow the protocols for the widely-used pre-processed subset VG150 \cite{XuZCF17}, which contains the most frequent 150 entities and 50 predicates. The dataset contains approximately 108k images, with
70\% for training and 30\% for testing. 
\textbf{2) Panoptic Scene Graph (PSG)} \cite{PSG-YangAGZZ022}. Filtered from COCO \cite{COCO-LinMBHPRDZ14} and VG datasets \cite{VG-KrishnaZGJHKCKL17}, the PSG dataset contains 133 object classes, including things, stuff, and 56 relation classes. This dataset has 46k training images and 2k testing images with panoptic segmentation and scene graph annotation. We follow the same data-processing pipelines from \cite{PSG-YangAGZZ022}.

\vspace{-1mm}
\paragraph{Video:}

\textbf{1) Action Genome (AG)} \cite{AG-JiK0N20} annotates 234,253 frame scene graphs for sampled frames from around 10K videos, based on Charades dataset \cite{SigurdssonVWFLG16}. The annotations cover 35 object categories and 25 predicates. The overall predicates consist of three types of predicates: attention, spatial, and contracting.
\textbf{2) Panoptic Video Scene Graph (PVSG)} \cite{PVSG-YangPLGCL0ZZLL23} consists of 400 videos, including 289 third-person
videos from VidOR \cite{VidOR-shang2019annotating} and 111 egocentric videos from EpicKitchens \cite{EPIC-KITCHENS-abs-1804-02748} and Ego4D \cite{Ego4D-GraumanWBCFGH0L22}. 
Among the videos, 62 videos feature birthday celebrations, while 35 videos center around ceremonies, providing rich content for contextual logic and reasoning.

\vspace{-1mm}
\paragraph{3D:}

\textbf{3D Scene Graph (3DSG)} \cite{3DDSG-WaldDNT20} includes 1335 3D reconstructed indoor scenes, 528 classes of objects, and 39 types of predicates.

\vspace{-1mm}
\paragraph{Text:}

\textbf{FACUTAL} \cite{FACTUAL-LiCZQHLJT23} is derived from VG \cite{VG-KrishnaZGJHKCKL17} dataset, which includes 4,042 classes of objects, 1,607 types of predicates, and 40,369 instances.

\begin{figure*}[!t]
    \centering
    \includegraphics[width=0.99\linewidth]{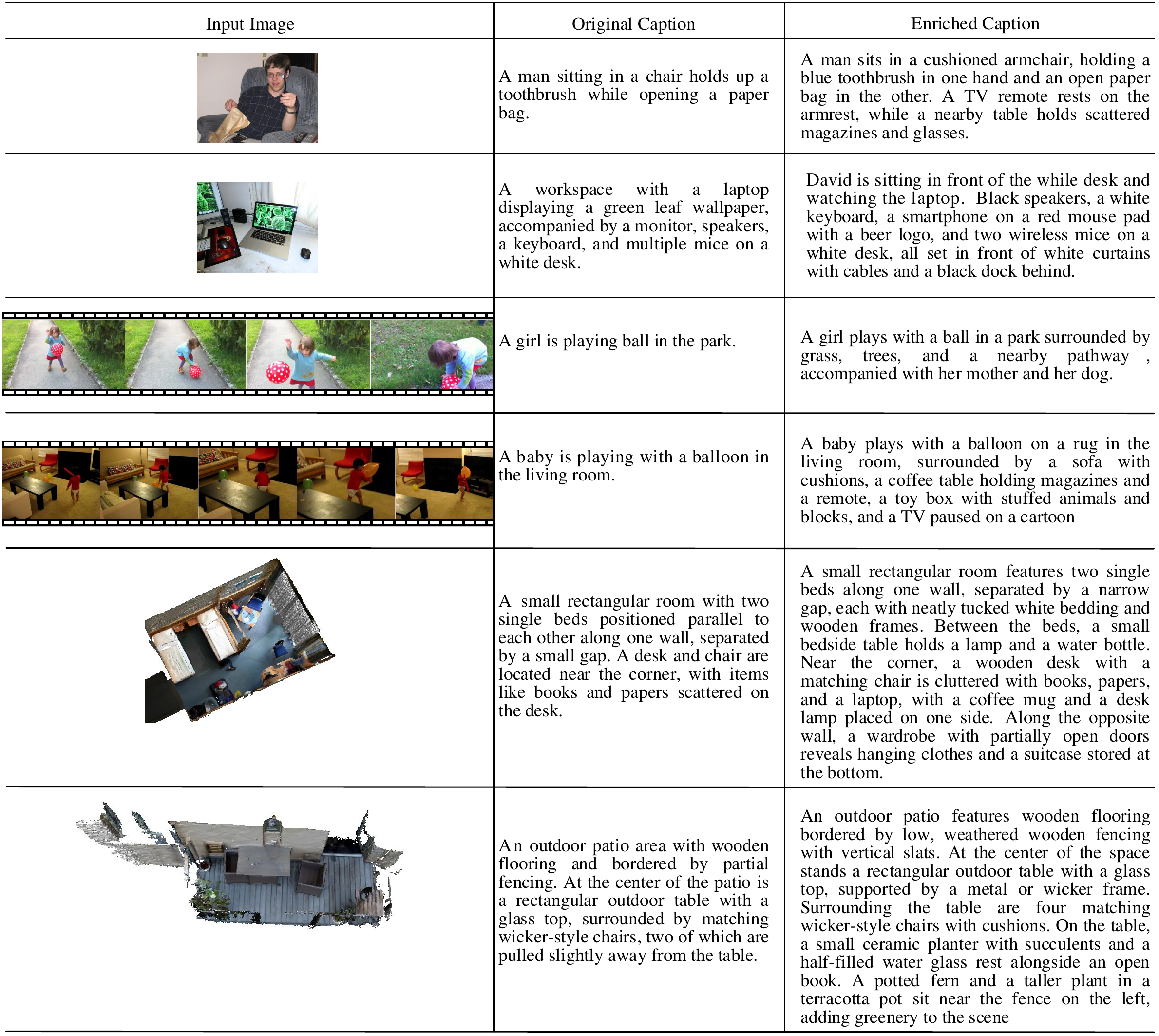}
    \vspace{-2mm}
    \caption{Examples of constructed Text-Image/Video/3D pair dataset with original caption and enriched caption. }
    \label{fig:enriched-caption}
    \vspace{-2mm}
\end{figure*}

\subsubsection{Multi-modal Dataset}
Here, we show the detailed process for constructing the USG dataset involving two input modalities.

\vspace{-1mm}
\paragraph{Text-Image ($\mathcal{S}-\mathcal{I}$).}
We leverage the three image caption datasets: COCO caption \cite{COCO-LinMBHPRDZ14},  Conceptual (CC) caption \cite{CC-SoricutDSG18}, and VG \cite{VG-KrishnaZGJHKCKL17} caption to build the Text-Image pair-wise SG. 
Specifically, following \cite{LLM4SGG-KimYJIM0P24}, we first employ the GPT-4o \cite{openai2023gpt4} to extract the triplets from the original caption.
Then, we align entities in the triplets with entity classes of interest and align predicates in the triplets with predicate classes
of interest. 
Finally, for the VG dataset with ISG annotation, we link the textual and visual objects through label matching.
For the COCO and CC datasets without annotated SG, we ground the extracted triplets over relevant image regions to get localized triplets via state-of-the-art grounding methods, i.e., Grounded SAM \cite{abs-2401-14159}.
Finally, we led to utilizing 64K images on the COCO caption dataset, 145K images on the CC caption dataset, and 57K images on the VG caption dataset.
Furthermore, we utilize GPT-4o to rephrase and enhance captions, aiming to increase the diversity and richness of textual descriptions, guided by the following prompts:
\begin{tcolorbox}[breakable, fontupper=\customfont, title=Rephrase and Enrich captions]
\vspace{-2mm}
{\small
\textbf{Input Data}: textual captions \\
\textbf{Instruction}: 
From the given sentence, the task is to enrich the caption with reasonable scenes. Let's take a few examples to understand how to enrich the captions.\\
\\
\textbf{[Example-1]}:\\
\textbf{Input}: A lady and a child near a park bench with kites and ducks flying in the sky and on the ground.\\
\textbf{Output}: A lady and a child near a park bench surrounded by lush greenery, with colorful kites soaring in the sky, ducks flying overhead, and a few waddling on the ground. Nearby, a serene pond reflects the vibrant scene, and children can be seen playing in the background.\\
\\
\textbf{[Example-2]}:\\
\textbf{Input}: Two men sit on a bench near the sidewalk and one of them talks on a cell phone.\\
\textbf{Output}: Two men sit on a wooden bench near a bustling sidewalk, shaded by nearby trees. One of them is engaged in a conversation on his cell phone, gesturing slightly with his free hand, while the other man sits calmly, gazing at passersby. \\
...
}
\end{tcolorbox}
We present two examples in the first two rows of Fig. \ref{fig:enriched-caption}.
After generating the enriched captions, we apply the aforementioned method used for the original captions to produce the final pairwise text-image SG annotations.

\vspace{-2mm}
\paragraph{Text-Video ($\mathcal{S}-\mathcal{V}$).}

To construct the text-video pairwise USG dataset, we select 400 videos from ActivityNet \cite{ActivityNet-HeilbronEGN15}, which includes dense caption annotations. 
Following the procedure for text-image pairs, we first extract triplets and align the entities and predicates with relevant concepts. 
Finally, we track textual objects in the videos by integrating frame-level object grounding results using Grounded SAM \cite{abs-2401-14159}. 
Additionally, we enrich the video captions to create textual SGs, incorporating partially nonliteral associations with the video content.
We depict two examples in the middle two rows of Fig. \ref{fig:enriched-caption}.

\vspace{-2mm}
\paragraph{Text-3D ($\mathcal{S}-\mathcal{D}$).}

To construct the text-3D pairwise USG dataset, we use the ScanRefer \cite{ScanRefer-ChenCN20} dataset, which contains 46,173 descriptions of 724 object types across 800 ScanNet \cite{ScanNet-DaiCSHFN17} scenes. 
Triplets are extracted from these descriptions using GPT-4o. 
Since ScanRefer provides object localizations, we directly align textual entities with 3D objects to establish associations between text and 3D data. 
Additionally, we enrich the textual descriptions to create partially overlapping text-3D USG datasets, enhancing diversity and coverage.
Two examples are demonstrated in the last two rows of  Fig. \ref{fig:enriched-caption}.

\vspace{-2mm}
\paragraph{Image-Video ($\mathcal{I}-\mathcal{V}$).}

To construct the image-video pairwise USG dataset, we utilize the existing PVSG \cite{PVSG-YangPLGCL0ZZLL23} video dataset. Specifically, we select the first frame of each video to construct frame-level ISGs. 
We then extract temporally non-adjacent video segments as the corresponding pairwise video. 
The associations between image objects and video objects are derived from the original PVSG annotations, ensuring accurate cross-modal connections.

\begin{figure}
    \centering
    \includegraphics[width=0.99\linewidth]{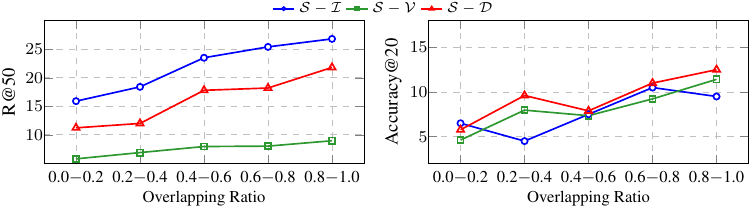}
    \caption{Performance of SGDet and association accuracy scores under varying overlap ratios between the two modalities.}
    \label{fig:overlapping-ratio}
\vspace{-2mm}
\end{figure}

\vspace{-2mm}
\paragraph{Image-3D ($\mathcal{I}-\mathcal{D}$).}
To construct the Image-3D USG dataset, we leverage the existing 3DSG \cite{3DDSG-WaldDNT20} dataset. 
Specifically, we randomly select 2D image views corresponding to 3D scenes. 
Using object annotations from the 3DSG dataset, we ground the objects in the selected images to obtain their positional information. 
The relationships among the detected objects are then derived from the original 3DSG annotations, resulting in complete SG annotations for the 2D image views.
The associations between objects in the image and 3D scenes are determined by whether the 3D objects can be successfully grounded in the image. By integrating ISGs, 3DSGs, and association relations, we construct the final Image-3D pairwise USG dataset.

\vspace{-1mm}

\subsection{Implementations}
\label{app:implementation}
\vspace{-1mm}

We initialize the text and image encoders using OpenCLIP~\cite{Radford2021LearningTV}, where the specific version of the image encoder is ConvNext-L.
we design the pixel decoder by following the approach in \cite{mask2former-ChengMSKG22,LiY0DWZLCL24}. 
For the point encoder, we adopt Point-BERT~\cite{Point-BERT-YuTR00L22} as the initialization, and for the point decoder, inspired by \cite{Point-BERT-YuTR00L22}, we implement a hierarchical propagation strategy with distance-based interpolation. 
After the encoding, all the features are projected into a 256-dimension using a linear layer. 
The mask decoder follows the design in ~\cite{mask2former-ChengMSKG22}.
We set the number of predefined learnable queries to 100.
The number of layers $L^{mask}$ is set as 9, with 3 transformer layers per scale.
The object associator is implemented as a 3-layer CNN with a kernel size of $3 \times 3$. 
For the two-way relation-aware object/subject module, the number of layers, $L_{RPC}$, is set to 4.
The relation decoder comprises a 6-layer transformer with an embedding dimension of 256.
During training, we used the AdamW optimizer with an initial learning rate of $10e-4$ and a weight decay of $10e-4$.
For the object detection loss weights, we set the $\lambda_{ce}=5.0$ and $\lambda_{dice}=5.0$, and $\lambda_{cls}=2.0$ for predictions matched with ground truth and 0.1 for the ``no object''.
In the final loss, we set the loss weights $\alpha=1.0, \beta=1.0$ and $\beta=0.8$, $\eta=0.6$.

\vspace{-1mm}

\section{Extended Experimental Results}
\label{app:experiments}
\vspace{-1mm}

We exhibit more experimental results here.

\vspace{-3mm}
\paragraph{The Impact of the Overlap Ratio.}

We delve into the analysis of the overlap ratio and its crucial role in influencing the performance of USG-Par. 
To this end, the $\mathcal{S}-\mathcal{I}$, $\mathcal{S}-\mathcal{I}$, and $\mathcal{S}-\mathcal{I}$ datasets are divided into five groups with overlap ratios ranging from 0.0 to 1.0. 
The results, presented in Fig.~\ref{fig:overlapping-ratio}, indicate that as the overlap ratio increases, the model achieves more accurate SG generation. 
This improvement can be attributed to the increased similarity between the two modalities, where complementary information enhances SG recognition performance. 
Furthermore, a higher overlap ratio allows the model to establish more precise associations between objects across modalities.

\begin{table}[!t]
  \centering
\fontsize{8}{10}\selectfont
\setlength{\tabcolsep}{1.2mm}
\begin{tabular}{lcccccccc}
\hline
$\mathcal{S}$ & $\mathcal{I}$ & $\mathcal{V}$ & $\mathcal{D}$ & $\mathcal{S}-\mathcal{I}$  & $\mathcal{S}-\mathcal{V}$ & $\mathcal{S}-\mathcal{D}$ & $\mathcal{I}-\mathcal{V}$ & $\mathcal{I}-\mathcal{D}$  \\
\hline
28.4 & 36.9 & 5.3  & 38.7 & 21.6   & 14.5 &  7.9  & 10.7 &  18.7  \\
25.1 & $\times$ & $\times$  & $\times$ & -  & - &  -  & - &  -  \\
27.6 & 34.0 & $\times$ & $\times$ & 16.0   & - &  -  & - & -    \\
28.1 & 34.9 & 4.9  & $\times$ & 19.2   & 14.0 &  -   & - & - \\
\hline
\end{tabular}%
\vspace{-2mm}
\caption{
The ablation of modality unification.
mR@20 scores are reported.
}
\vspace{-4mm}
\label{tab:modality_unification}
\end{table}

\vspace{-3mm}
\paragraph{The Effect of Modality Unification.}
Tab. \ref{tab:modality_unification} compares the performance of our method across different numbers of unified modalities. The results demonstrate that incorporating additional modalities consistently improves performance across all datasets. 
This observation highlights the complementary nature of information across modalities when representing a scene. 
By unifying multiple modalities, our model effectively leverages this complementary information, resulting in enhanced performance compared to using a single modality alone.

\begin{table}[!t]
  \centering
\fontsize{8}{11}\selectfont
\setlength{\tabcolsep}{5mm}
\begin{tabular}{cc}
\hline
$\mathcal{I}-\mathcal{V}$ & $\mathcal{I}-\mathcal{D}$ \\
\midrule
\multicolumn{2}{l}{$\bullet$ \textit{Only Corresponding Supervision.}} \\
25.1 & 18.4  \\
\cdashline{1-2}
\multicolumn{2}{l}{$\bullet$ \textit{No Supervision.} }\\
23.4  & 16.8 \\
\hline
\end{tabular}%
\vspace{-2mm}
\caption{
Emergent zero-shot on the $\mathcal{I}-\mathcal{V}/\mathcal{D}$ dataset, where the model is only trained on $\mathcal{S}-\mathcal{I}/\mathcal{V}/\mathcal{D}$ data. 
}
\vspace{-4mm}
\label{tab:emergent}
\end{table}

\vspace{-3mm}
\paragraph{The Emergent Capability of USG-Par.}

Tab. \ref{tab:emergent} presents the emergent zero-shot SG generation performance on $\mathcal{I}-\mathcal{V}$ and $\mathcal{I}-\mathcal{D}$, where USG-Par is trained solely on $\mathcal{S}-\mathcal{I}/\mathcal{V}/\mathcal{D}$. 
The results indicate that our model achieves performance comparable to supervised learning approaches. 
This demonstrates the strong capability of USG-Par to align image, video, and 3D modalities effectively, leveraging text as a unifying bridge.

\vspace{-3mm}
\paragraph{More Visualizations.}

\begin{figure}[ht]
    \centering
    \includegraphics[width=0.99\linewidth]{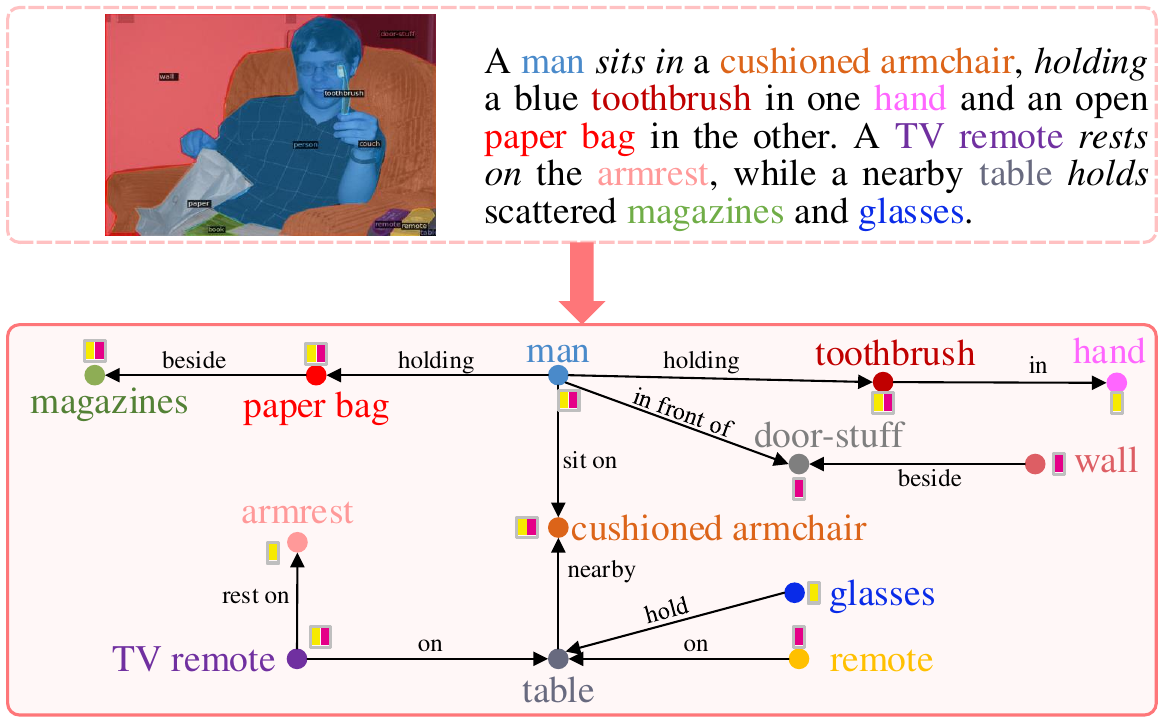}
    \vspace{-2mm}
    \caption{Visualization of USG generated from text and image}
    \label{fig:app-t-i-case}
    \vspace{-1mm}
\end{figure}

\begin{figure}[ht]
    \centering
    \includegraphics[width=0.99\linewidth]{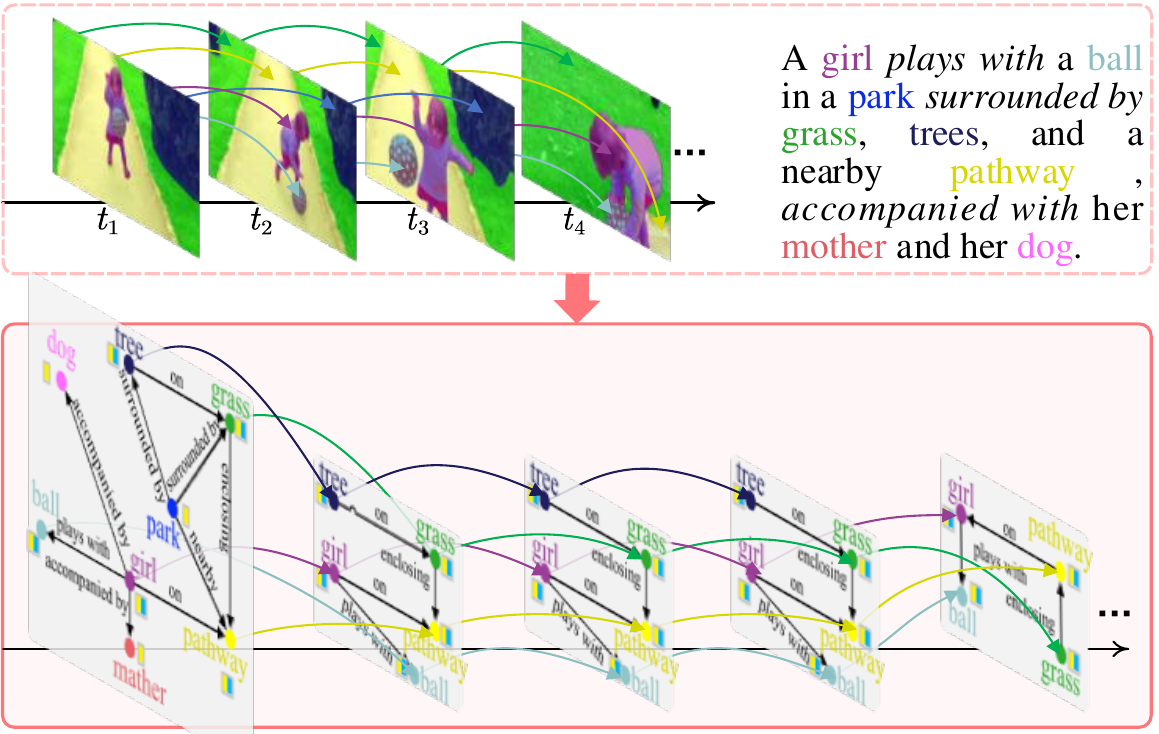}
    \vspace{-2mm}
    \caption{Visualization of USG generated from text and video}
    \label{fig:t-v-case}
    \vspace{-1mm}
\end{figure}

\begin{figure}[ht]
    \centering
    \includegraphics[width=0.99\linewidth]{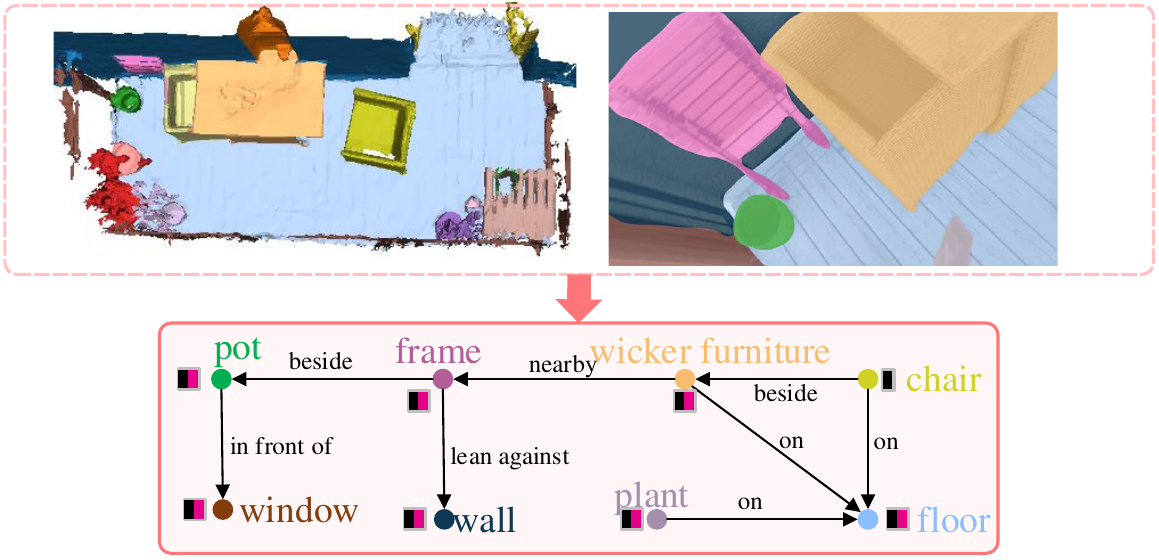}
    \vspace{-2mm}
    \caption{Visualization of USG generated from 3D and the corresponding view image.}
    \label{fig:i-3d-case}
    \vspace{-1mm}
\end{figure}

Here, we provide more visualizations of generated USG  from various input modality combinations, including 1) text + image in Fig. \ref{fig:app-t-i-case}, 2) text + video in Fig. \ref{fig:t-v-case}, and 2) image + 3D in Fig. \ref{fig:i-3d-case}.

\end{document}